%% file: main.tex
\documentclass[10pt,twocolumn,letterpaper]{article}

\usepackage{wacv}              

\input{preamble}

\definecolor{wacvblue}{rgb}{0.21,0.49,0.74}
\usepackage[pagebackref,breaklinks,colorlinks,allcolors=wacvblue]{hyperref}

\def\wacvPaperID{167} 
\def\confName{WACV}
\def\confYear{2027}

\title{Depth-Guided Contrastive Learning for \\2D Representations with 3D Spatial Awareness}

\author{Liang Zeng\\
KU Leuven\\
Belgium\\
{\tt\small liang.zeng@kuleuven.be}
\and
Maarten Vergauwen\\
KU Leuven\\
Belgium\\
{\tt\small maarten.vergauwen@kuleuven.be}
}

\begin{document}
\maketitle
\input{sec/0_abstract}

\input{sec/1_intro}

\input{sec/2_related_work}
\input{sec/3_method}
\input{sec/4_experiment}
\input{sec/5_ablation}

\input{sec/6_visualization}

{
    \small
    \bibliographystyle{ieeenat_fullname}
    \bibliography{main}
}

\input{sec/7_sup}

\end{document}

%% file: preamble.tex
\usepackage[dvipsnames]{xcolor}
\usepackage{amsmath}
\usepackage{amssymb}
\usepackage{booktabs}
\usepackage{float}
\usepackage{makecell}
\usepackage{multirow}
\usepackage{multicol}
\usepackage{algorithm}
\usepackage{algorithmic}

\usepackage[table]{xcolor}
\definecolor{lightgreen}{RGB}{217, 234, 211}

\usepackage{graphicx}
\usepackage{float}      

%% file: sec/0_abstract.tex
\begin{abstract}
Standard contrastive learning frameworks are mainly designed from a semantic perspective, yet learning 2D visual representations that preserve 3D spatial structure is also important for scene understanding. In this work, we propose Depth-Guided Contrastive Learning (DGCL), a simple auxiliary objective that injects 3D spatial awareness into 2D contrastive representation learning. Our key idea is to use depth to convert local 3D proximity into contrastive similarity: pixels that are closer in 3D space are encouraged to have more similar representations than pixels that are farther apart. Instead of relying on absolute depth values, DGCL formulates supervision through relative 3D distance comparisons among randomly sampled pixels, making the objective invariant to depth scale, efficient to compute, and easy to integrate into existing contrastive frameworks. Experiments across different datasets and models show that DGCL consistently improves 2D representation learning and benefits semantic downstream tasks by stronger spatial and geometric understanding. The code is available on {\footnotesize \url{https://github.com/LeungTsang/DGCL}.}
\end{abstract}

%% file: sec/1_intro.tex
\section{Introduction}
\label{sec:introduction}

Self-supervised learning has become a central paradigm for learning transferable visual representations from large-scale unlabeled data~\cite{DBLP:conf/iccv/WangG15,DBLP:conf/iccv/DoerschGE15,DBLP:conf/eccv/NorooziF16,DBLP:conf/eccv/ZhangIE16,DBLP:conf/cvpr/PathakGDDH17,DBLP:conf/iclr/GidarisSK18}, including recently-introduced contrastive learning~\cite{DBLP:conf/icml/ChenK0H20, DBLP:conf/cvpr/He0WXG20}, and masked image modeling~\cite{DBLP:conf/cvpr/HeCXLDG22, DBLP:conf/cvpr/Xie00LBYD022}. Among different self-supervised objectives, contrastive learning remains a simple and effective framework by pulling positive samples closer in the embedding space while pushing negative samples apart~\cite{DBLP:conf/cvpr/HadsellCL06,DBLP:conf/icml/ChenK0H20, DBLP:conf/cvpr/He0WXG20}.

For contrastive learning, the key question is how to define meaningful positive and negative relationships. In traditional image-level contrastive learning, positive pairs are commonly generated from different augmentations of the same image, while other images are treated as negatives~\cite{DBLP:conf/cvpr/WuXYL18, DBLP:conf/icml/ChenK0H20,DBLP:conf/cvpr/He0WXG20, DBLP:conf/nips/GrillSATRBDPGAP20}. This simple assumption has proven effective for semantic representation learning, but it is less suitable for dense prediction tasks~\cite{DBLP:conf/eccv/LinMBHPRDZ14, DBLP:journals/ijcv/EveringhamGWWZ10, DBLP:conf/cvpr/CordtsORREBFRS16, DBLP:journals/ijcv/ZhouZPXFBT19}, where meaningful visual information often lies in local structures and relationships between pixels or regions. Pixel-level contrastive learning addresses this issue by contrasting local features rather than only global image embeddings~\cite{DBLP:conf/nips/PinheiroABGC20, DBLP:conf/cvpr/WangZSKL21, DBLP:conf/cvpr/XieL00L021, DBLP:conf/nips/BardesPL22}. However, it also introduces a more challenging problem: how should the fine-grained similarity between pixels be defined?

Existing methods usually define pixel-level similarity from a semantic perspective. Some approaches rely on supervised or unsupervised object discovery and segmentation masks to relate pixels that are likely to belong to the same object or semantic category~\cite{DBLP:conf/nips/ZhangM20, DBLP:conf/nips/ZhangTRR21, DBLP:conf/cvpr/WangZSKL21, DBLP:conf/cvpr/XieL00L021, DBLP:conf/nips/WenZZZQ22, DBLP:conf/cvpr/StegmullerLBTT23}. While effective, supervised priors reduce scalability, and unsupervised grouping methods can be sensitive to low-level appearance cues such as color, texture, and illumination. Other methods estimate pixel similarity from the learned feature space itself~\cite{DBLP:conf/cvpr/WangZSKL21, DBLP:conf/cvpr/XieL00L021, DBLP:conf/nips/WenZZZQ22}, assuming that the learned representations are self-consistent. These approaches avoid external supervision, but their learned relationships may not necessarily correspond to semantics or physical structures.

\begin{figure*}
  \centering
  \includegraphics[width=0.89\linewidth]{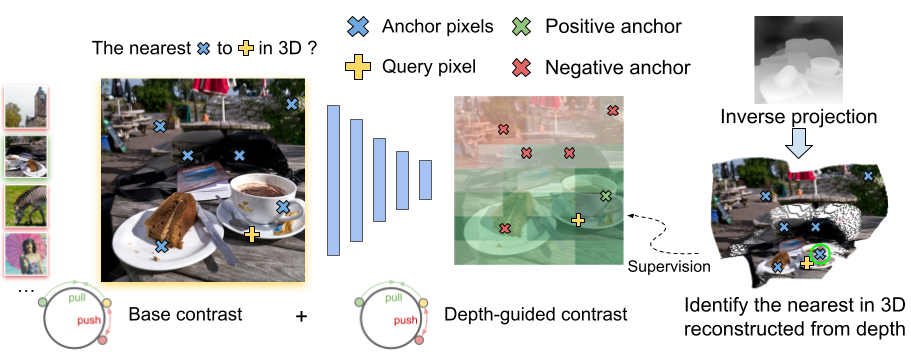}
  \label{img}
  \caption{Overview of Depth-Guided Contrastive Learning (DGCL). In addition to the base contrastive objective, DGCL introduces an auxiliary pixel-level contrastive task guided by depth. Given a query pixel and randomly sampled anchor pixels, the anchor closest to the query in the 3D space reconstructed from depth is selected as the positive, while the others are treated as negatives. This nearest-anchor prediction objective converts relative 3D proximity into contrastive supervision, encouraging 2D features to capture local 3D structure.}
  \label{method}
\end{figure*}

In this work, we propose to define pixel-level similarity from a geometric perspective instead of treating pixel similarity as purely semantic. We does NOT assume that pixels close in 3D space necessarily share the same semantic label, nor use 3D proximity to approximate specific semantic classes. Our assumption is weaker: pixels that are close in 3D space are often physically or structurally related. They may lie on the same surface, belong to nearby interacting objects, or participate in local scene structures. Encouraging their representations to be correlated can therefore help the model learn 3D-aware local organization of the scene. Since 3D structure and semantic understanding are closely related in dense prediction tasks~\cite{DBLP:conf/cvpr/HoyerDCKSG21}, learning representations that better reflect the underlying 3D structure also benefits semantic downstream tasks.

To this end, we introduce Depth-Guided Contrastive Learning (DGCL), an auxiliary objective that injects 3D spatial awareness into 2D contrastive representation learning. Given an RGB image and its depth map, DGCL randomly samples pixels as \textbf{anchors} and compares their relative 3D distances to each \textbf{query} pixel. For each query, the closest anchor in 3D is selected as the positive, while the remaining anchors are treated as negatives. This pretext task provides a complementary geometric signal, encouraging feature similarities to reflect 3D structure while the base contrastive objective preserves semantic invariance.

DGCL is simple and versatile. First and foremost, DGCL relies on relative distance comparisons among sampled pixels within an image, instead of formulating a fixed transformation from absolute 3D distances to similarity scores, making the objective robust to depth scale. Secondly, DGCL make use of increasingly accessible depth from RGB-D sensors, synthetic environments, stereo systems, and especially modern monocular depth estimators~\cite{DBLP:journals/pami/RanftlLHSK22, DBLP:conf/cvpr/YangKHXFZ24} to enhance 2D visual representation learning, rather than using more expensive curated 3D datasets, point clouds, or camera trajectories. Finally, unlike multi-modal contrastive learning that aligns RGB and depth embeddings, DGCL directly constructs geometric supervision for RGB features without additional depth encoders. As a standalone module, it can be added to existing contrastive frameworks with minimal architectural modification.

We summarize our contributions as follows:

\begin{itemize}
\item We propose Depth-Guided Contrastive Learning, a contrastive objective that uses relative 3D proximity to inject geometric awareness into 2D contrastive frameworks.

\item We show that DGCL consistently improves transfer performance across different contrastive frameworks, pretraining datasets, and downstream tasks.
\end{itemize}

%% file: sec/2_related_work.tex
\section{Related Work}
\label{sec:relatedwork}

\subsection{Contrastive learning}

Self-supervised visual representation learning has achieved strong transfer performance by exploiting supervisory signals derived from data itself. Among different self-supervised paradigms, contrastive learning is a foundational family of methods that learns an embedding space by pulling positive samples together and pushing negative samples apart~\cite{DBLP:conf/cvpr/HadsellCL06, DBLP:conf/icml/ChenK0H20,DBLP:conf/nips/ChenKSNH20,DBLP:conf/cvpr/He0WXG20,DBLP:journals/corr/abs-2003-04297}. Early methods typically define positive pairs as different augmentations of the same image and treat other images as negatives~\cite{DBLP:conf/icml/ChenK0H20}, while later variants improve the objective through momentum encoders~\cite{DBLP:conf/cvpr/He0WXG20}, clustering~\cite{DBLP:conf/nips/CaronMMGBJ20, DBLP:conf/iclr/0001ZXH21}, asymmetric architectures~\cite{DBLP:conf/nips/GrillSATRBDPGAP20, DBLP:conf/cvpr/ChenH21}, redundancy reduction~\cite{DBLP:conf/iclr/BardesPL22}, or information maximization~\cite{DBLP:conf/icml/ZbontarJMLD21}. These methods have proven effective for image-level recognition, where the instance-discrimination assumption provides a simple and scalable learning signal.

Image-level contrastive learning is not always optimal for dense prediction tasks such as object detection~\cite{DBLP:conf/eccv/LinMBHPRDZ14} and semantic segmentation~\cite{DBLP:journals/ijcv/EveringhamGWWZ10, DBLP:conf/cvpr/CordtsORREBFRS16, DBLP:journals/ijcv/ZhouZPXFBT19}, where local features and spatial correspondences are critical. To address this issue, many works extend contrastive learning to pixel-level~\cite{DBLP:conf/nips/PinheiroABGC20, DBLP:conf/nips/ZhangM20, DBLP:conf/cvpr/WangZSKL21, DBLP:conf/nips/ZhangTRR21, DBLP:conf/cvpr/XieL00L021, DBLP:conf/nips/BardesPL22, lebailly2023cribo}, patch-level~\cite{DBLP:conf/iccv/XieDWZXSL021, DBLP:conf/cvpr/BaiCKYB22, DBLP:conf/eccv/Ci00O22}, region-or object-level~\cite{DBLP:conf/nips/XieZLOL21, DBLP:conf/cvpr/YangWZL21, DBLP:conf/cvpr/RohSKK21, DBLP:conf/eccv/XuZ0T22, DBLP:conf/eccv/WangWWYS22, DBLP:conf/nips/WenZZZQ22, DBLP:conf/cvpr/LiZZYLZWYZHCG22, DBLP:conf/cvpr/GeMKL023, DBLP:conf/cvpr/StegmullerLBTT23} representation learning. In dense contrastive learning, the definition of the local similarity is a key design choice. Some methods derive similarity internally from the learned feature space itself, for example through online clustering~\cite{DBLP:conf/nips/WenZZZQ22, DBLP:conf/cvpr/StegmullerLBTT23}, nearest-neighbor matching~\cite{DBLP:conf/cvpr/WangZSKL21, DBLP:conf/nips/BardesPL22}, or feature propagation~\cite{DBLP:conf/cvpr/XieL00L021}. Although these approaches avoid external supervision, the learned relationships may inherit appearance-level biases from RGB images, such as texture, color, dataset bias, or shortcut cues~\cite{DBLP:journals/natmi/GeirhosJMZBBW20,DBLP:conf/nips/RobinsonSYBJS21}. 
Other methods introduce external priors, such as object proposals, segmentation masks, or hierarchical regions generated by supervised or unsupervised grouping models~\cite{DBLP:conf/nips/ZhangM20,DBLP:conf/iccv/HenaffKAOVC21,DBLP:conf/nips/WeiGWHL21}. 
These priors can provide more structured pixel relationships, but they may require additional models, preprocessing, task-specific assumptions, or supervised annotations.

Our method belongs to the family of pixel-level contrastive learning, but differs from prior work in how pixel similarity is defined. Instead of relying on semantic grouping or RGB-based feature similarity, we use depth to construct a geometric notion of similarity. The key assumption is that pixels close in 3D space are likely to be physically related and should therefore have correlated representations. In this way, DGCL introduces 3D spatial awareness into 2D contrastive representation learning while remaining a lightweight auxiliary objective that can be combined with existing contrastive frameworks.

\subsection{2D representation learning with 3D information}

Learning 2D representations with 3D information has attracted increasing attention, as many visual tasks require not only semantic recognition but also spatial and geometric understanding. A common direction is to use RGB-D data, multi-view images, camera poses, point clouds, or reconstructed geometry to inject 3D priors into image encoders~\cite{DBLP:conf/iccv/HouXGDN21, DBLP:journals/corr/abs-2012-13089, DBLP:conf/cvpr/ChenCPLW22}. Such methods exploit correspondences across different views of the same scene, treating image regions that correspond to the same 3D point or region as positive pairs. Although effective, the requirement of structured data limits the broader applicability. Another line of work incorporates 3D information through multimodal or multitask pretraining. Multimodal contrastive learning can align RGB and depth representations in a shared latent space~\cite{DBLP:conf/eccv/TianKI20, DBLP:conf/cvpr/GirdharELSAJM23}, while masked image modeling approaches can train image encoders by jointly reconstructing depth, surface normals, or other geometric modalities~\cite{DBLP:conf/eccv/BachmannMAZ22, DBLP:conf/cvpr/HouDHDN23}. However, these methods are often computationally expensive with additional networks to process different modalities and the learning of 3D awareness is mostly implicit. 

Depth has also been used more directly in dense contrastive learning. PixDepth~\cite{DBLP:conf/wacv/SaadPKDCF23}, for instance, improves pixel-level contrastive learning by using depth as an external cue to filter unlikely positive pairs in PixPro~\cite{DBLP:conf/cvpr/XieL00L021}. This is closely related to our motivation, but its formulation remains tied to a specific pixel-level contrastive framework and mainly uses depth to refine positive-pair selection. In contrast, DGCL formulates depth guidance as a standalone contrastive objective based on relative 3D proximity. Rather than thresholding absolute depth differences or aligning RGB features with a depth encoder, DGCL compares sampled pixels in 3D space and selects positives according to their relative closeness.

Compared with prior 3D-guided representation learning methods, DGCL has three distinctive properties. First, it only requires paired RGB images and depth maps, without camera poses, multi-view sequences, point clouds, or reconstructed scenes. Second, it uses relative 3D distance comparisons without relying on absolute metric depth, making it robust to different sources and scales of depth. Third, DGCL directly supervises 2D visual features with geometric relationships derived from depth, instead of treating depth as a reconstruction target or aligning RGB features with a separate depth embedding. Therefore, DGCL provides a simple and flexible way to inject 3D spatial awareness into 2D contrastive representation learning.

%% file: sec/3_method.tex
\section{Methodology}
\label{sec:methodology}

\subsection{Back-Projection from Depth to 3D}

DGCL uses depth to construct a geometric relationship between pixels. Given a 2D pixel and its depth value, we first back-project it to an approximate 3D coordinate using the pinhole camera model. Let $(u,v)$ denote the pixel coordinate, $d(u,v)$ its corresponding depth value, and $\mathbf{K}$ the camera intrinsic matrix. The 3D coordinate $\mathbf{p}(u,v)=[x,y,z]^{\top}$ is computed as
\begin{equation}
\mathbf{p}(u,v)
= d(u,v)\mathbf{K}^{-1}
\begin{bmatrix}
u \\ v \\ 1
\end{bmatrix}.
\label{eq:inverse_projection_general}
\end{equation}

In many pretraining datasets, accurate camera intrinsics are unavailable. Thus, we use a simplified intrinsic matrix by setting the skew to zero, the aspect ratio to one, and the principal point to the image center. When the focal length is unknown, we use a fixed hypothesized focal length $f$ as a hyperparameter. For an image of width $w$ and height $h$, the simplified intrinsic matrix is
\begin{equation}
\mathbf{K} =
\begin{bmatrix}
f & 0 & w/2 \\
0 & f & h/2 \\
0 & 0 & 1
\end{bmatrix}.
\label{eq:simplified_intrinsic}
\end{equation}

The back-projected 3D coordinates can be written as
\begin{equation}
\mathbf{p}(u,v)
=
\begin{bmatrix}
d(u,v)\frac{u-w/2}{f} \\
d(u,v)\frac{v-h/2}{f} \\
d(u,v)
\end{bmatrix}.
\label{eq:inverse_projection}
\end{equation}

The resulting 3D coordinates are not intended to provide a precise metric reconstruction of the scene. They serve as a geometric cue for comparing the relative 3D distances between sampled pixels. Moreover, DGCL is applied to feature maps with lower spatial resolution than the input image and therefore coarse 3D reconstruction is sufficient. The simplified back-projection makes the proposed objective applicable even when depth is estimated, scale-ambiguous, or obtained from different sources.

\subsection{From 3D Proximity to Contrastive Similarity}

Given the back-projected 3D coordinates, DGCL converts 3D proximity into a geometric contrastive learning signal. A direct way would be to transform the absolute 3D distance between two pixels into a continuous similarity score. However, this is unreliable in practice. For example, two adjacent pixels on a distant background surface may have a larger 3D distance than two adjacent pixels on a nearby foreground object, although both pairs may represent locally coherent regions. In addition to the scene layout, the magnitude of 3D distance depends on the sources of depth with different camera intrinsics. Therefore, using a hand-crafted transformation can introduce undesirable scale dependence.

To avoid the ambiguous direct transformation from 3D distance to feature similarity, DGCL uses relative 3D distance comparisons as the learning objective. For each image, we randomly sample $N$ groups of $K$ anchor pixels each from the set of all pixels $\Omega$. Let $\mathbf{r}_{i}\in\mathbb{R}^{C}$ denote the normalized representation of query pixel $i$, and let $\mathbf{p}_{i}\in\mathbb{R}^{3}$ denote its back-projected 3D coordinate. For the $n$-th anchor group, we denote the anchor pixels as $\mathcal{A}_{n}=\{a_{n,1},\ldots,a_{n,K}\}$. The positive anchor for a query pixel $i$ is selected as the anchor with the smallest 3D distance to the query:
\begin{equation}
y_{i,n}
=
\arg\min_{k\in\{1,\ldots,K\}}
\left\|
\mathbf{p}_{i}-\mathbf{p}_{a_{n,k}}
\right\|_{2}.
\label{eq:nearest_anchor}
\end{equation}
The remaining anchors in the same group are treated as negatives. The contrastive logits are computed from the feature similarity between the query and anchor representations:
\begin{equation}
s_{i,n,k}
=
\frac{
\mathbf{r}_{i}^{\top}\mathbf{r}_{a_{n,k}}
}{\tau},
\label{eq:dgcl_logits}
\end{equation}
where $\tau$ is the temperature. To use all pixels within an image $\Omega$ as query pixels and average $N$ groups, the overall depth-guided contrastive loss is then defined as
\begin{equation}
\mathcal{L}_{\mathrm{DGCL}}
=
\frac{1}{N|\Omega|}
\sum_{n=1}^{N}
\sum_{i\in\Omega}
-
\log
\frac{
\exp(s_{i,n,y_{i,n}})
}{
\sum_{k=1}^{K}\exp(s_{i,n,k})
},
\label{eq:dgcl_loss}
\end{equation}
The pseudo code is provided in Algorithm ~\ref{code}.

\renewcommand{\algorithmicrequire}{\textbf{Input:}}
\renewcommand{\algorithmicensure}{\textbf{Output:}} 
\begin{algorithm}
    \begin{algorithmic}
        \REQUIRE Dense representation $\textbf{R}(b,c,h,w)$, 3D coordinates $\textbf{P}(b,3,h,w)$ 
        
        \ENSURE DGCL loss $\textbf{L}$
        \STATE \textcolor[RGB]{0,200,0}{\# randomly sample $N\times K$ anchor pixels for each image}
        \STATE $ \textbf{A} = empty(size=(b, N, K, 2)).uniform\_(-1, 1) $
        \STATE $ \textbf{R}_{a} = grid\_sample(\textbf{R}, grid=\textbf{A}, mode="nearest") $
        \STATE $ \textbf{P}_{a} = grid\_sample(\textbf{P}, grid=\textbf{A}, mode="nearest") $
        \STATE \textcolor[RGB]{0,200,0}{\# find the nearest anchor for each pixel}
        \STATE $\textbf{Y} = norm(\textbf{P}.view(b, 3, 1, h, w) - \textbf{P}_{a}.view(b, 3, N\times K, 1, 1), dim=1)$
        \STATE $\textbf{targets} = argmin(\textbf{Y}.view(b, N, K, h, w), dim=2)$ 
        \STATE \textcolor[RGB]{0,200,0}{\# nearest as positive, compute the InfoNCE loss}
        \STATE $\textbf{logits} = einsum('bchw ,bcnk \rightarrow bnhwk', (\textbf{R}, \textbf{R}_{a}))/\tau$
        \STATE $\textbf{L} = cross\_entropy(\textbf{logits}, \textbf{targets})$

        \RETURN $\textbf{L}$
        
    \end{algorithmic} 
    \caption{Depth-guided contrastive loss in PyTorch style} 
    \label{code} 
\end{algorithm}

This formulation transforms 3D proximity into learnable feature similarity and the model learns to discriminate pixels according to their relative 3D distances. Our method does not rely on an absolute distance threshold, because the target is determined by the relative order of distances within each randomly sampled anchor group. Therefore, multiplying all 3D coordinates by a positive scale factor does not change the selected positive anchor. As a result, DGCL is robust to the metric scale of depth and can be applied to sensor depth, synthetic depth, or monocularly estimated depth.

Finally, DGCL is optimized jointly with a base contrastive learning objective:
\begin{equation}
\mathcal{L}
=
\mathcal{L}_{\mathrm{base}}
+
\lambda \mathcal{L}_{\mathrm{DGCL}},
\label{eq:overall_loss}
\end{equation}
where $\lambda$ is the contribution of the depth-guided objective.

\subsection{Distribution of positive samples}
\begin{figure}[H]
  \centering
  \begin{minipage}[t]{0.19\linewidth}
    \centering
    \includegraphics[width=\linewidth]{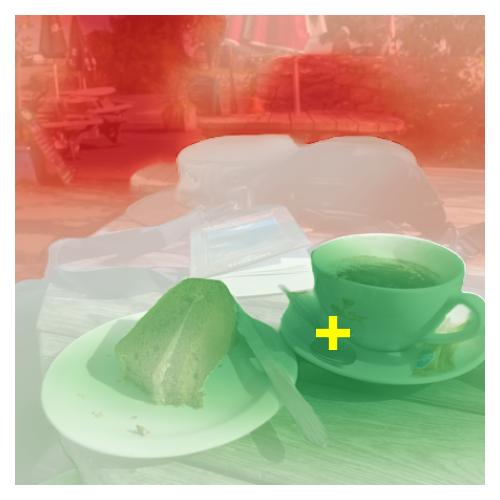}\\
    {$K=2$}
  \end{minipage}
  \begin{minipage}[t]{0.19\linewidth}
    \centering
    \includegraphics[width=\linewidth]{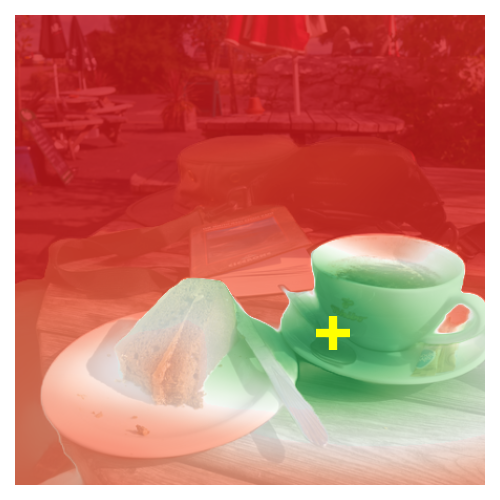}\\
    {$K=4$}
  \end{minipage}
  \begin{minipage}[t]{0.19\linewidth}
    \centering
    \includegraphics[width=\linewidth]{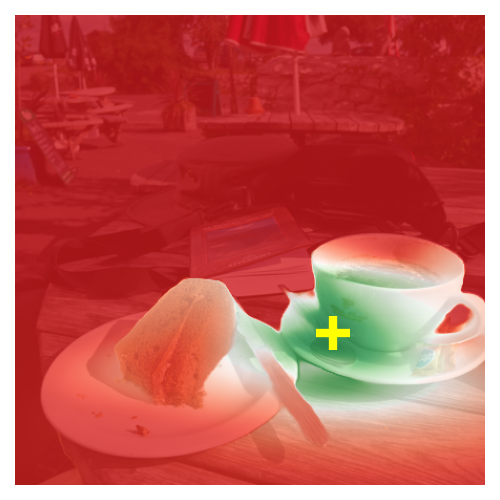}\\
    {$K=8$}
  \end{minipage}
  \begin{minipage}[t]{0.19\linewidth}
    \centering
    \includegraphics[width=\linewidth]{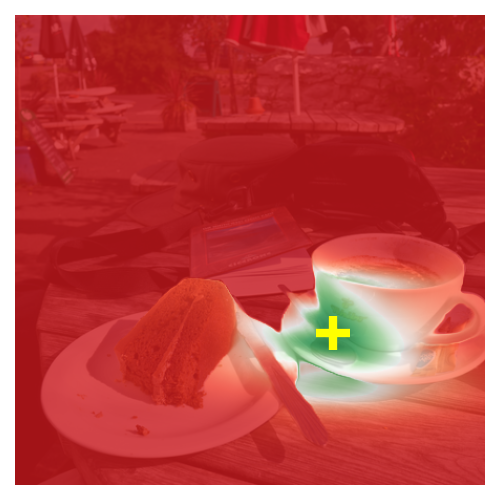}\\
    {$K=16$}
  \end{minipage}
  \begin{minipage}[t]{0.19\linewidth}
    \centering
    \includegraphics[width=\linewidth]{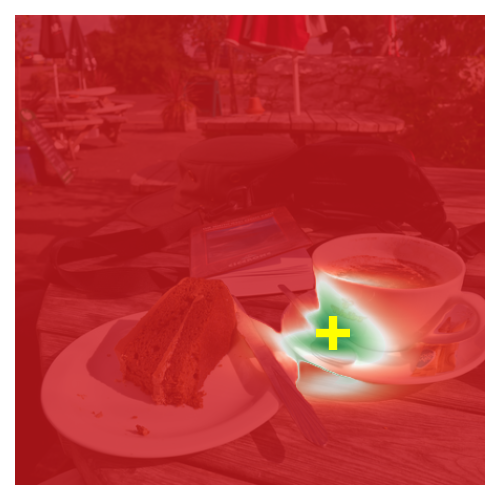}\\
    {$K=32$}
  \end{minipage}
  \caption{The probability distribution of positive samples around the query pixel \textbf{\textcolor{Goldenrod}{+}} with different numbers of anchors $K$. Green and red indicate high and low probability, respectively.}
  \label{fig:k}
\end{figure}

The repeated random sampling provides a soft notion of geometric similarity. A pixel that is closer to the query in 3D is more likely to become the nearest anchor across repeated trials. Assume there is a pixel $a_i$ and it is the $i$-th closest pixel to the query pixel in 3D, where $i=1$ indicates the closest. The probability that $a_i$ is the positive sample to the query pixel is given by Eq.(~\ref{p}), which means $a_i$ is selected as an anchor and all other anchor pixels are farther. As the Figure ~\ref{fig:k} shows, the numbers of anchors $K$ controls the concentration of positive sample around the query pixel.


\begin{equation}
\begin{aligned}
P(a_i\ is\ positive) = \frac{K}{n}(\frac{n-i+1}{n})^{K-1}
\end{aligned}
\label{p}
\end{equation}


Although the closed-form probability is available, we stick to the Monte Carlo sampling in practice, to avoid the expensive computation of pairwise distance between all valid pixels.

%% file: sec/4_experiment.tex
\section{Experiments}
\label{experiment}
We perform extensive experiments to examine the effectiveness of DGCL in multiple scenarios. We integrate DGCL into MoCo v2~\cite{DBLP:journals/corr/abs-2003-04297}, MoCo v3~\cite{DBLP:conf/iccv/ChenXH21} and SlotCon~\cite{DBLP:conf/nips/WenZZZQ22}, referred to as DGCL-M2, DGCL-M3, and DGCL-S, respectively, to study its interaction with different based methods in different scenarios. MoCo v2 and MoCo v3 are classic image-level contrastive learning methods using Resnet~\cite{He_2016_CVPR} and Vision Transformers~\cite{dosovitskiy2021an}, respectively, and DGCL supplements it with a pixel-level learning objective. SlotCon features data-driven pixel grouping by prototypes and DGCL is expected to further improve its dense representation learning. The implementation is based on their official PyTorch codes. We compare DGCL with the immediate baselines to isolate the benefit of DGCL and other strong baselines to position the performance in the field. We report the average results of 3 independent runs and the standard deviations are provided in supplementary materials ~\ref{standard deviation}. 

For DGCL, we set the number of anchors to $K=6$, the number of repeated anchor samplings to $N=64$, and the temperature in Eq.~\ref{eq:dgcl_logits} to $\tau=0.1$. Depth maps are estimated using MiDaS~\cite{DBLP:journals/pami/RanftlLHSK22} with the $v3.1~Swin2_{L-384}$ model and we use its suggested hypothesized focal length of $f=900$.

\subsection{Pre-training on object-centric ImageNet}
\label{experiment_im}
\textbf{Implementation details}
We train a ResNet-50~\cite{He_2016_CVPR} for 200 epochs and keep the training settings of the original base methods. We evaluate two variants: DGCL-M2, built on MoCo v2, and DGCL-S, built on SlotCon. 

\noindent\textbf{Evaluation protocol} We follow the evaluation protocol of SlotCon~\cite{DBLP:conf/nips/WenZZZQ22} and report transfer performance on COCO object detection and instance segmentation~\cite{DBLP:conf/eccv/LinMBHPRDZ14}, as well as semantic segmentation on Pascal VOC~\cite{DBLP:journals/ijcv/EveringhamGWWZ10}, Cityscapes~\cite{DBLP:conf/cvpr/CordtsORREBFRS16}, and ADE20K~\cite{DBLP:journals/ijcv/ZhouZPXFBT19}. Additionally, we evaluate models on COCO keypoint detection. COCO tasks are evaluated with Detectron2~\cite{wu2019detectron2} while others use MMSegmentation~\cite{mmseg2020}.

\noindent\textbf{Transfer learning results} Table~\ref{im_result} reports the transfer learning results after ImageNet pretraining. DGCL consistently improves its corresponding base method. Compared with the 200-epoch MoCo v2 baseline, DGCL-M2 improves COCO detection AP from \textbf{39.2} to \textbf{40.2}, mask AP from \textbf{36.1} to \textbf{37.1}, and keypoint AP from \textbf{66.0} to \textbf{66.6}. The improvement is also clear on semantic segmentation, especially on Pascal VOC, where mIoU increases from \textbf{72.4} to \textbf{76.2}. DGCL also improves the stronger SlotCon baseline. DGCL-S increases COCO detection AP from \textbf{41.8} to \textbf{42.1}, mask AP from \textbf{37.8} to \textbf{38.2}, and keypoint AP from \textbf{66.7} to \textbf{67.2}. Larger gains are observed on semantic segmentation, with improvements of \textbf{0.7}, \textbf{1.9}, and \textbf{1.5} mIoU on Cityscapes, Pascal VOC, and ADE20K, respectively. Notably, with only 200 epochs of pretraining, DGCL-S reaches comparable or better performance than DetCon trained for 1000 epochs. These results suggest that depth-guided local geometric supervision complements semantic grouping and improves the transferability of dense visual representations.

\begin{table*}\small
  \centering
  \begin{tabular}{@{}l|c|c|c|c|c|c|c @{}}
    \toprule
    \multirow{2}*{Method} & \multirow{2}*{Epoch}  & COCO Detection  & COCO Segmentation & COCO Keypoint  & City. & VOC & ADE \\
     & & $\text{AP}^\text{b}$ / $\text{AP}_{50}^\text{b}$ / $\text{AP}_\text{75}^\text{b}$ & $\text{AP}^\text{m}$ / $\text{AP}_\text{50}^\text{m}$ / $\text{AP}_\text{75}^\text{m}$ & $\text{AP}^\text{k}$ / $\text{AP}_\text{50}^\text{k}$ / $\text{AP}_\text{75}^\text{k}$ & mIoU & mIoU & mIoU \\
    \midrule
    MoCo v2~\cite{DBLP:journals/corr/abs-2003-04297} & 800 & 40.4 / 60.1 / 44.2 & 36.5 / 57.2 / 39.2 & 66.8 / 87.4 / 72.4 & 76.2 & 73.7 & 36.9 \\
    BYOL~\cite{DBLP:conf/nips/GrillSATRBDPGAP20} & 1000 & 41.1 / 61.8 / 44.6 & 37.6 / 59.3 / 40.6 & 66.4 / 87.4 / 72.4 & 75.0 & 72.1 & 37.6 \\
    DenseCL~\cite{DBLP:conf/cvpr/WangZSKL21} & 200 & 40.3 / 59.9 / 44.3 & 36.4 / 57.0 / 39.2 & 66.8 / 87.2 / 72.9 & 76.2 & 72.8 & 38.1\\
    PixPro~\cite{DBLP:conf/cvpr/XieL00L021} & 400 & 40.4 / 60.4 / 44.1 & 37.0 / 57.7 / 39.9 & 67.1 / \textbf{87.8} / 73.4 & 76.8 & 73.4 & 38.8\\
    DetCon~\cite{DBLP:conf/iccv/HenaffKAOVC21} & 1000 & 41.6 / 62.3 / 45.4 & \textbf{38.2} / \textbf{59.9} / \textbf{41.2} & 66.7 / \textbf{87.8} / 72.4 & 76.7 & 76.3 & 39.9\\
    \midrule
    MoCo v2~\cite{DBLP:journals/corr/abs-2003-04297} & 200 & 39.2 / 59.3 / 42.8 & 36.1 / 56.8 / 38.7 & 66.0 / 86.9 / 71.8 & 76.1 & 72.4 & 37.3\\
    \rowcolor{lightgreen}
    DGCL-M2 & 200 & 40.2 / 60.4 / 44.1 & 37.1 / 58.0 / 40.1 & 66.6 / 87.6 / 72.5 & 76.4 & 76.2 & 38.7\\
    \midrule
    SlotCon~\cite{DBLP:conf/nips/WenZZZQ22} & 200 & 41.8 / 62.2 / 45.7 & 37.8 / 59.1 / 40.7 & 66.7 / 87.5 / 72.8 & 76.3 & 75.0 & 38.8\\
    \rowcolor{lightgreen}
    DGCL-S & 200 & \textbf{42.1} / \textbf{62.4} / \textbf{46.2} & \textbf{38.2} / 59.6 / 40.9 & \textbf{67.2} / \textbf{87.8} / \textbf{73.5} & \textbf{77.0} & \textbf{76.9} & \textbf{40.3}\\
    \bottomrule
  \end{tabular}
  \caption{Transfer learning results after ImageNet pretraining. DGCL consistently improves its corresponding base method across COCO object detection, instance segmentation, keypoint detection, and semantic segmentation. DGCL-S, trained for only 200 epochs, achieves performance comparable to or better than the 1000-epoch DetCon baseline on most metrics.}
  \label{im_result}
\end{table*}

\subsection{Pre-training on scene-based COCO}
\label{experiment_coco}
\textbf{Implementation details} Following SlotCon~\cite{DBLP:conf/nips/WenZZZQ22}, we pretrain ResNet-50 on the COCO train2017 split~\cite{DBLP:conf/eccv/LinMBHPRDZ14} for 800 epochs. DGCL-M2 and DGCL-S use the same base training settings as their corresponding methods. The DGCL hyperparameters are kept unchanged.

\noindent\textbf{Evaluation protocol} We reuse the transfer evaluation protocol of SlotCon~\cite{DBLP:conf/nips/WenZZZQ22}, including COCO object detection, keypoint detection and instance segmentation~\cite{DBLP:conf/eccv/LinMBHPRDZ14}, as well as semantic segmentation on Pascal VOC~\cite{DBLP:journals/ijcv/EveringhamGWWZ10}, Cityscapes~\cite{DBLP:conf/cvpr/CordtsORREBFRS16}, and ADE20K~\cite{DBLP:journals/ijcv/ZhouZPXFBT19}.

\noindent\textbf{Transfer learning results} Table~\ref{coco_result} shows the transfer learning results after COCO pretraining. Compared with MoCo v2, DGCL-M2 brings substantial gains across COCO dense prediction and semantic segmentation tasks. It improves COCO detection AP from \textbf{38.5} to \textbf{40.1}, mask AP from \textbf{34.8} to \textbf{36.9}, and keypoint AP from \textbf{65.8} to \textbf{66.7}. The gains are even more pronounced for semantic segmentation, with improvements of \textbf{2.6}, \textbf{3.2}, and \textbf{2.7} mIoU on Cityscapes, Pascal VOC, and ADE20K, respectively. DGCL-S further improves the stronger SlotCon baseline. It increases COCO detection AP from \textbf{41.0} to \textbf{41.3}, mask AP from \textbf{37.0} to \textbf{37.9}, and keypoint AP from \textbf{66.3} to \textbf{67.4}. It also improves semantic segmentation by \textbf{0.5}, \textbf{1.1}, and \textbf{0.9} mIoU on Cityscapes, Pascal VOC, and ADE20K, respectively. DGCL-S achieves the best performance across all reported metrics. Compared with ImageNet pretraining, the gains from COCO pretraining are more consistent and larger, suggesting that depth-guided contrastive learning is particularly beneficial when pretraining on scene-centric images that contain richer spatial layout and geometric structure.

\begin{table*}\small
  \centering
   
  \begin{tabular}{@{}l|c|c|c|c|c|c|c @{}}
    \toprule
    \multirow{2}*{Method} & \multirow{2}*{Epoch}  & COCO Detection  & COCO Segmentation & COCO Keypoint  & City. & VOC & ADE \\
     & & $\text{AP}^\text{b}$ / $\text{AP}_{50}^\text{b}$ / $\text{AP}_\text{75}^\text{b}$ & $\text{AP}^\text{m}$ / $\text{AP}_\text{50}^\text{m}$ / $\text{AP}_\text{75}^\text{m}$ & $\text{AP}^\text{k}$ / $\text{AP}_\text{50}^\text{k}$ / $\text{AP}_\text{75}^\text{k}$ & mIoU & mIoU & mIoU \\
    \midrule
    DenseCL~\cite{DBLP:conf/cvpr/WangZSKL21} & 800 & 39.6 / 59.3 / 43.3 & 35.7 / 56.5 / 38.4 & 66.3 / 87.1 / 72.2 & 75.8 & 71.6 & 37.1\\
    PixPro~\cite{DBLP:conf/cvpr/XieL00L021} & 800 & 40.5 / 60.5 / 44.0 & 36.6 / 57.8 / 39.0 & 66.4 / 87.3 / 72.4 & 75.2 & 72.0 & 38.3\\
    DetCon~\cite{DBLP:conf/iccv/HenaffKAOVC21} & 1000 & 39.8 / 59.5 / 43.5 & 35.9 / 56.4 / 38.7 &  66.1 / 87.2 / 71.8 & 76.1 & 70.2 & 38.1\\
    ORL~\cite{DBLP:conf/nips/XieZLOL21} & 800& 40.3 / 60.2 / 44.4 & 36.3 / 57.3 / 38.9 & 66.6 / 87.0 / 72.7 & 75.6 & 70.9 & 36.7\\
    \midrule
    MoCo v2~\cite{DBLP:journals/corr/abs-2003-04297} & 800& 38.5 / 58.1 / 42.1 & 34.8 / 55.3 / 37.3 & 65.8 / 86.9 / 71.6 & 73.8 & 69.2 & 36.2\\
    \rowcolor{lightgreen}
    DGCL-M2 & 800 & 40.1 / 60.3 / 43.6 & 36.9 / 57.9 / 39.8 & 66.7 / 87.5 / 72.5 & 76.4 & 72.4 & 38.9\\
    \midrule
    SlotCon~\cite{DBLP:conf/nips/WenZZZQ22} & 800 & 41.0 / 61.1 / 45.0 & 37.0 / 58.3 / 39.8 & 66.3 / 87.4 / 73.1 & 76.2 & 71.6 & 39.0\\
    \rowcolor{lightgreen}
    DGCL-S & 800 & \textbf{41.3} / \textbf{61.6} / \textbf{45.1} & \textbf{37.9} / \textbf{59.1} / \textbf{40.8} & \textbf{67.4} / \textbf{87.8} / \textbf{73.9} & \textbf{76.7} & \textbf{72.7} & \textbf{39.9}\\
    \bottomrule
  \end{tabular}
  \caption{Transfer learning results after COCO pretraining. DGCL consistently improves both MoCo v2 and SlotCon across COCO object detection, instance segmentation, keypoint detection, and semantic segmentation, indicating that depth-guided geometric supervision benefits pretraining on scene images. Moreover, DGCL-S achieves the best performance on all metrics.}
  \label{coco_result}
\end{table*}

\subsection{Pre-training Vision Transformers}

\textbf{Implementation details} Following MoCo v3~\cite{DBLP:conf/iccv/ChenXH21}, we train ViT-Small with patch size 16 for 300 epochs on ImageNet. 

\noindent\textbf{Evaluation protocol} We evaluate the pre-trained models on COCO instance segmentation and Pascal VOC and ADE20k semantic segmentation. For COCO instance segmentation, we leverage ViTDet implemented by detectron2 with the default configuration and a batch size of 4. For semantic segmentation, we train UperNet with ViT backbone implemented by MMSegmentation~\cite{mmseg2020} on Pascal VOC for 20k iterations and ADE20K for 40k iterations.

\noindent\textbf{Transfer learning results}
Table~\ref{vit_im} reports the transfer learning results and DGCL consistently improves MoCo v3. On COCO instance segmentation, DGCL-M3 improves mask AP from \textbf{33.7} to \textbf{34.9}, with gains of \textbf{1.6} $\text{AP}_{50}^{m}$ and \textbf{1.5} $\text{AP}_{75}^{m}$. The improvement also transfers to semantic segmentation, where DGCL-M3 improves mIoU from \textbf{79.1} to \textbf{79.9} on Pascal VOC and from \textbf{39.4} to \textbf{40.7} on ADE20K. Compared with DINO~\cite{Caron_2021_ICCV}, DGCL-M3 obtains better performance on Pascal VOC but remains lower on COCO instance segmentation and ADE20K. This is expected, as DINO uses a different self-distillation framework and is pretrained for a longer schedule. The main comparison here is therefore between MoCo v3 and DGCL-M3 under the same contrastive pretraining framework. These results show that DGCL is not limited to convolutional networks and can also improve ViT-based contrastive representation learning.

\begin{table}[H]\small
  \centering
   
  \begin{tabular}{@{}l|c|c|c|c @{}}
    \toprule
    \multirow{2}*{Method} & \multirow{2}*{Epoch}  & COCO Seg. & VOC & ADE \\
    & &$\text{AP}^\text{m}$ / $\text{AP}_\text{50}^\text{m}$ / $\text{AP}_\text{75}^\text{m}$& mIoU & mIoU\\
    \midrule
    DINO~\cite{Caron_2021_ICCV}  & 400 & \textbf{37.1} / \textbf{59.3} / \textbf{39.4} & 77.9 & \textbf{42.0} \\
    \midrule
    MoCo v3~\cite{DBLP:conf/iccv/ChenXH21} & 300  & 33.7 / 54.4 / 35.6 & 79.1 & 39.4 \\
    \rowcolor{lightgreen}
    DGCL-M3 &  300 & 34.9 / 56.0 / 37.1 & \textbf{79.9} & 40.7 \\
    
    \bottomrule
  \end{tabular}
  \caption{Transfer learning results of ViT-S/16 models pretrained on ImageNet. DGCL-M3 consistently improves MoCo v3 under the same pretraining setting, showing that DGCL can also benefit transformer-based backbones.}
  \label{vit_im}
\end{table}

%% file: sec/5_ablation.tex
\section{Ablation and Analysis}
\label{ablation}

We conduct further experiments to analyze how DGCL compares with alternative ways of using depth, pseudo-mask guidance, and depth of different quality. Other hyperparameters are analyzed in supplementary materials~\ref{hyperparameter}. Unless otherwise specified, we use MoCo v2 as the base contrastive framework and pretrain all models on COCO for 200 epochs with the same training setting.

\subsection{Comparison with other methods that use depth}

We first compare DGCL with two alternative ways of incorporating depth into contrastive pretraining. The first baseline adds depth regression as an auxiliary pretext task, following the common practice of using depth prediction to encourage geometric understanding~\cite{DBLP:conf/iccv/HouXGDN21} by multi-task learning. The second baseline follows contrastive multiview coding (CMC)~\cite{DBLP:conf/eccv/TianKI20}, where RGB and depth are processed as different modalities and aligned in a shared embedding space. For a fair comparison, all methods are built on MoCo v2 and pretrained on COCO for 200 epochs. We evaluate transfer performance on COCO instance segmentation and semantic segmentation on Cityscapes and Pascal VOC.

Table~\ref{other_depth} reports the results. All depth-based objectives improve over the MoCo v2 baseline, confirming that depth provides useful geometric supervision for semantic representation learning. Among them, DGCL achieves the strongest performance across all metrics. The comparison suggests that how depth is used is important. Depth regression encourages the model to predict depth values directly, which may emphasize low-level local cues needed for depth prediction but less useful for semantic understanding. CMC aligns RGB and depth embeddings, but requires an additional depth branch and learns geometric information only implicitly through cross-modal representation matching. In contrast, DGCL provides explicit geometric supervision for representations without a depth encoder for embedding or a decoder for regression.

\begin{table}[H]
  \centering
   
  \begin{tabular}{@{}l|c|c|c @{}}
    \toprule
    \multirow{2}*{Method} & COCO Segmentation  & City. & VOC \\
    &$\text{AP}^\text{m}$ / $\text{AP}_\text{50}^\text{m}$ / $\text{AP}_\text{75}^\text{m}$ & mIoU & mIoU\\
    \midrule
    MoCo v2 & 33.4 / 53.6 / 35.8 & 72.6 & 65.3\\
    \midrule
    +CMC~\cite{DBLP:conf/eccv/TianKI20}  & 35.1 / 55.3 / 37.5 & 75.6 & 69.6 \\
    +Regression & 34.8 / 55.1 / 37.2 & 74.8 & 68.6 \\
    \rowcolor{lightgreen}
    +DGCL  & \textbf{35.5} / \textbf{56.2} / \textbf{38.0} & \textbf{76.5} & \textbf{70.7}\\
    \bottomrule
  \end{tabular}
  \caption{Comparison with alternative ways of using depth. All depth-based objectives improve the baseline, while DGCL achieves the best transfer performance without an additional depth encoder for  embedding or decoder for regression.}
  \label{other_depth}
\end{table}

\subsection{Comparison with segmentation mask guidance}

\begin{figure}[H]
  \centering
  \begin{minipage}{0.8\linewidth}
    \centering

    \begin{minipage}[t]{0.19\linewidth}
      \centering
      \includegraphics[width=\linewidth]{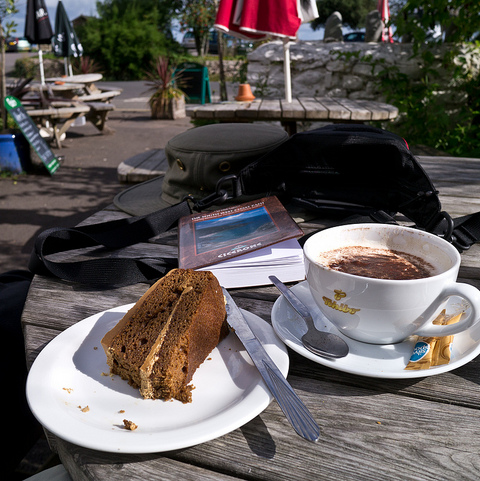}\\[-0.2em]
      {\scriptsize Image}
    \end{minipage}
    \hfill
    \begin{minipage}[t]{0.19\linewidth}
      \centering
      \includegraphics[width=\linewidth]{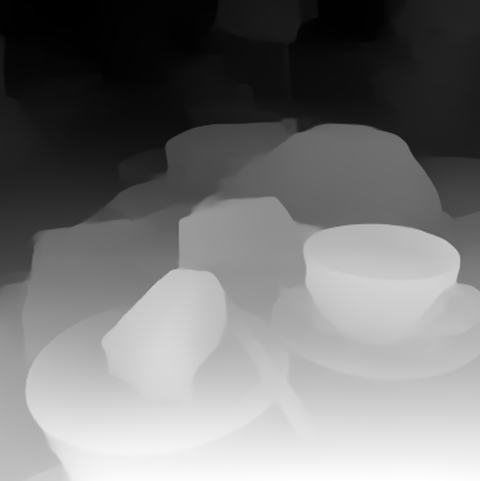}\\[-0.2em]
      {\scriptsize Depth}
    \end{minipage}
    \hfill
    \begin{minipage}[t]{0.19\linewidth}
      \centering
      \includegraphics[width=\linewidth]{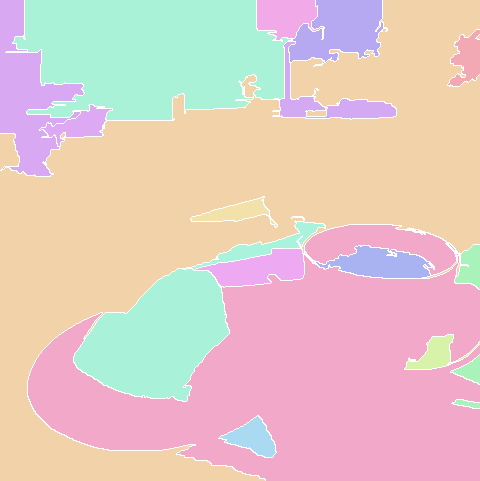}\\[-0.2em]
      {\scriptsize FH~\cite{DBLP:journals/ijcv/FelzenszwalbH04}}
    \end{minipage}
    \hfill
    \begin{minipage}[t]{0.19\linewidth}
      \centering
      \includegraphics[width=\linewidth]{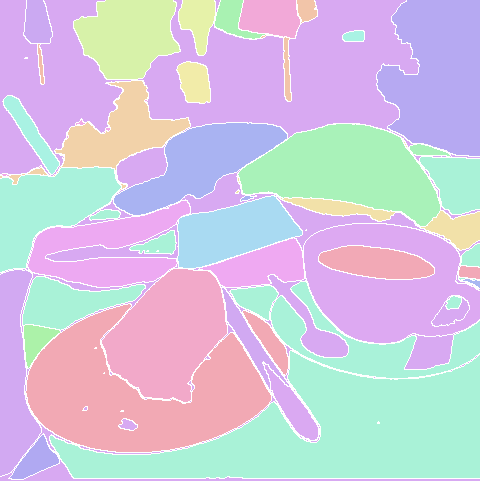}\\[-0.2em]
      {\scriptsize SAM~\cite{kirillov2023segany}}
    \end{minipage}
    \hfill
    \begin{minipage}[t]{0.19\linewidth}
      \centering
      \includegraphics[width=\linewidth]{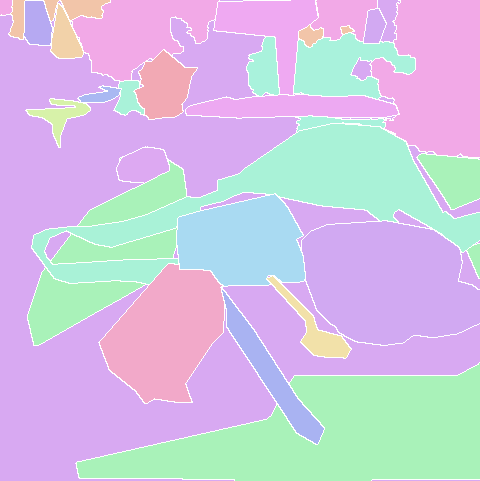}\\[-0.2em]
      {\scriptsize GT}
    \end{minipage}

  \end{minipage}

  \caption{Class-agnostic masks by FH~\cite{DBLP:journals/ijcv/FelzenszwalbH04}, SAM~\cite{kirillov2023segany}, and GT.
  }
  \label{fig:mask_example}
\end{figure}

We further compare DGCL with region-based guidance derived from segmentation masks. We consider masks generated by the Felzenszwalb-Huttenlocher (FH) segmentation algorithm~\cite{DBLP:journals/ijcv/FelzenszwalbH04}, masks generated by SAM~\cite{kirillov2023segany}, and class-agnostic ground-truth COCO instance masks, demonstrated in Figure ~\ref{fig:mask_example}. For each mask source, we compute a pooled region representation and use it as the positive target for pixels inside the corresponding region, forming an auxiliary pixel-level contrastive objective on top of MoCo v2. 

As shown in Table~\ref{other_guidance}, mask-based guidance consistently improves over MoCo v2, indicating that grouping pixels into object-like or region-like units is beneficial for dense representation learning. Stronger masks generally lead to better transfer performance: SAM outperforms FH, and ground-truth masks further improve the results while DGCL still achieves the best performance. 

\begin{table}[H]
  \centering
  \begin{tabular}{@{}l|c|c|c @{}}
    \toprule
    \multirow{2}*{Method} & COCO Segmentation  & City. & VOC \\
    &$\text{AP}^\text{m}$ / $\text{AP}_\text{50}^\text{m}$ / $\text{AP}_\text{75}^\text{m}$ & mIoU & mIoU\\
    \midrule
    MoCo v2 & 33.4 / 53.6 / 35.8 & 72.6 & 65.3\\
    \midrule
    +FH~\cite{DBLP:journals/ijcv/FelzenszwalbH04}  & 34.2 / 54.3 / 36.6 & 73.7 & 65.6 \\
    +SAM~\cite{kirillov2023segany}  & 34.8 / 55.0 / 37.2 & 75.8 & 67.7 \\
    +GT & 35.1 / 55.4 / 37.7  & 76.0 & 69.8 \\
    \rowcolor{lightgreen}
    +DGCL  & \textbf{35.5} / \textbf{56.2} / \textbf{38.0} & \textbf{76.5} & \textbf{70.7}\\
    \bottomrule
  \end{tabular}
  \caption{Comparison between segmentation mask guidance and depth-guided geometric guidance. Mask-based objectives improve MoCo v2 by grouping pixels within regions, while DGCL achieves stronger transfer performance by using relative 3D proximity to define pixel-level contrastive relationships.}
  \label{other_guidance}
\end{table}

\begin{figure}[H]

  \centering
  \begin{minipage}{1.0\linewidth}
  \centering

  \begin{minipage}[c]{0.08\linewidth}
    \centering
    \rotatebox[origin=c]{90}{\footnotesize Image}
  \end{minipage}
  \hfill
  \begin{minipage}[c]{0.90\linewidth}
    \centering
    \includegraphics[width=\linewidth]{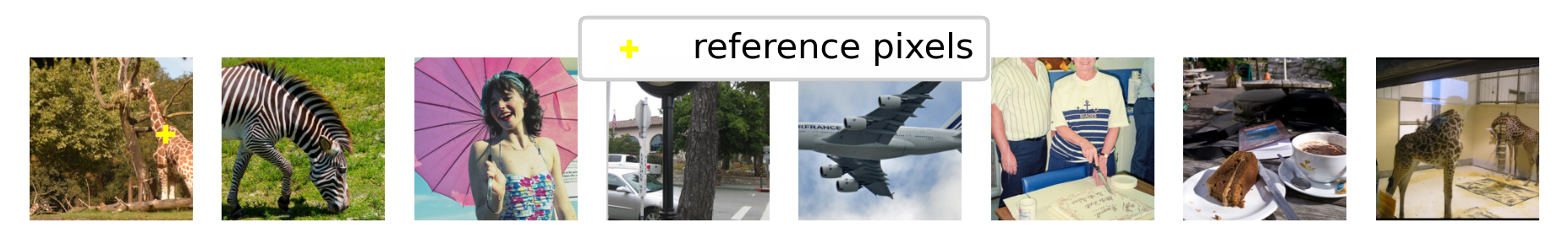}
  \end{minipage}

  \begin{minipage}[c]{0.08\linewidth}
    \centering
    \rotatebox[origin=c]{90}{\footnotesize Depth}
  \end{minipage}
  \hfill
  \begin{minipage}[c]{0.90\linewidth}
    \centering
    \includegraphics[width=\linewidth]{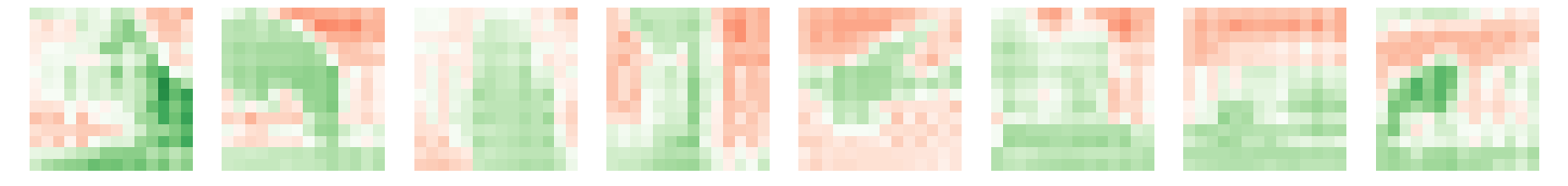}
  \end{minipage}

   \begin{minipage}[c]{0.08\linewidth}
    \centering
    \rotatebox[origin=c]{90}{\footnotesize GT}
  \end{minipage}
  \hfill
  \begin{minipage}[c]{0.90\linewidth}
    \centering
    \includegraphics[width=\linewidth]{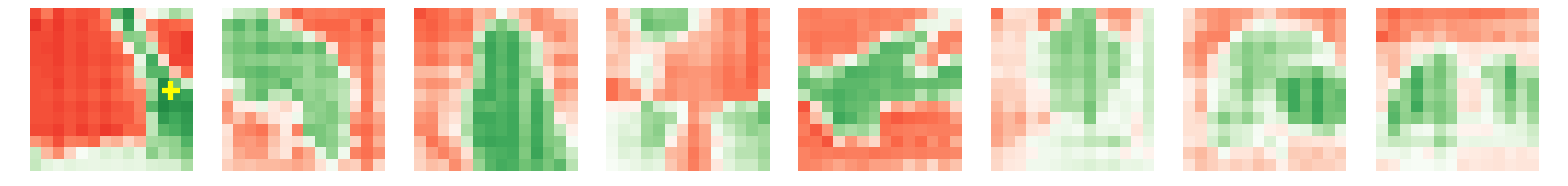}
  \end{minipage}
  \end{minipage}
  \caption{Pixel similarity with respect to the reference pixel marked by the yellow cross \textbf{\textcolor{Goldenrod}{+}} in the first image. Learning from either depth or class-agnostic GT tends to group the salient objects across different images regardless of their categories. However, DGCL may yield smoother geometry-aware similarity induced by relative 3D proximity for easier tuning in downstream tasks.}
  \label{fig:mask}
\end{figure}

This result highlights the difference between region-level grouping and geometric guidance and Figure~\ref{fig:mask} qualitatively compares the learned similarity maps. Mask-based objectives mainly encourage pixels within the same mask to have similar representations, but may not correctly relate different regions, especially across images. DGCL instead defines pixel relationships through relative 3D proximity, providing a complementary geometric signal that is not limited to object boundaries. Moreover, DGCL produces smoother feature maps than hard mask guidance, which may make the representations easier to adapt to downstream tasks. This helps explain why DGCL can outperform even mask guidance based on unlabelled ground-truth instance annotations.

\subsection{Sensitivity to Depth Quality}

\begin{figure}[H]
  \centering
  \begin{minipage}{1.0\linewidth}
    \centering

    \begin{minipage}[t]{0.15\linewidth}
      \centering
      \includegraphics[width=\linewidth]{fig/ablation_quality/img.png}\\[-0.2em]
      {\scriptsize Image}
    \end{minipage}
    \hfill
    \begin{minipage}[t]{0.15\linewidth}
      \centering
      \includegraphics[width=\linewidth]{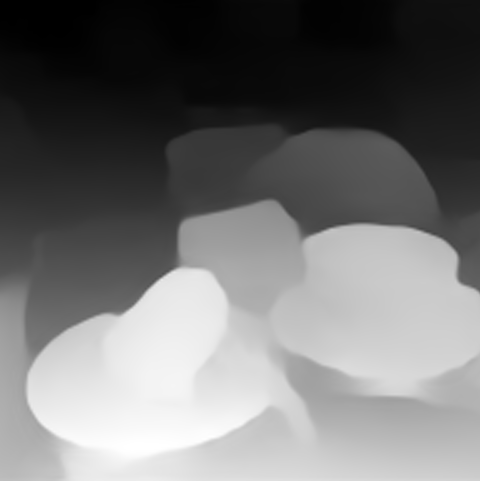}\\[-0.2em]
      {\scriptsize Small$_{256}$}
    \end{minipage}
    \hfill
    \begin{minipage}[t]{0.15\linewidth}
      \centering
      \includegraphics[width=\linewidth]{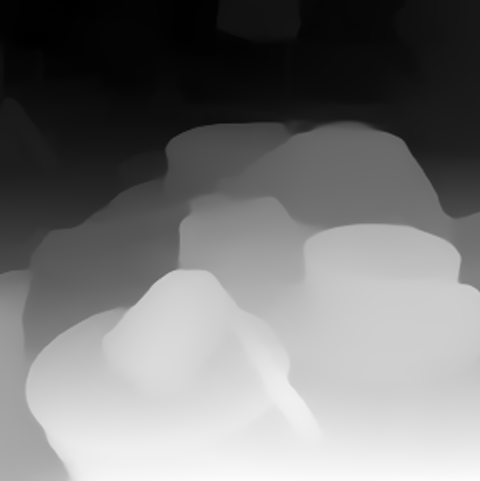}\\[-0.2em]
      {\scriptsize Large$_{384}$}
    \end{minipage}
    \hfill
    \begin{minipage}[t]{0.15\linewidth}
      \centering
      \includegraphics[width=\linewidth]{fig/ablation_quality/dpt_large_384.png}\\[-0.2em]
      {\scriptsize DPT$_{L-384}$}
    \end{minipage}
    \hfill
    \begin{minipage}[t]{0.15\linewidth}
      \centering
      \includegraphics[width=\linewidth]{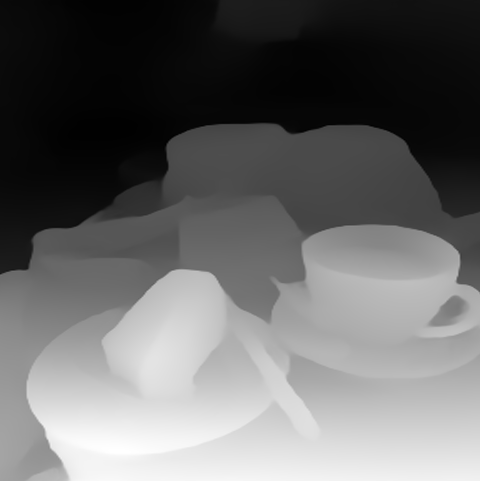}\\[-0.2em]
      {\scriptsize Swin2$_{L-384}$}
    \end{minipage}
    \hfill
    \begin{minipage}[t]{0.15\linewidth}
      \centering
      \includegraphics[width=\linewidth]{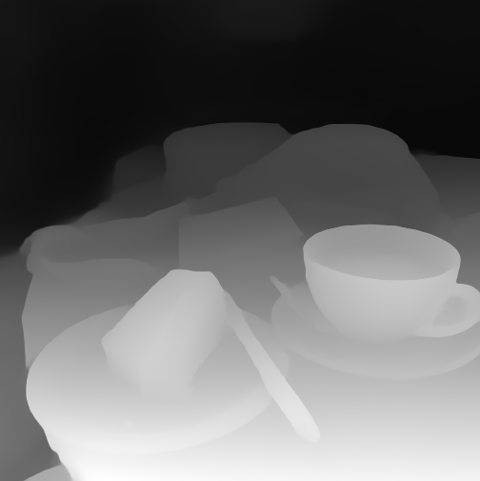}\\[-0.2em]
      {\scriptsize BEiT$_{L-512}$}
    \end{minipage}

  \end{minipage}

  \caption{Depth maps generated by different MiDaS models~\cite{DBLP:journals/pami/RanftlLHSK22}.
  }
  \label{fig:depth_example}
\end{figure}

We analyze how the quality of depth estimates affects DGCL. Specifically, we compare depth maps generated by different MiDaS models, whose relative depth quality is indicated by the MiDaS depth score~\cite{DBLP:journals/pami/RanftlLHSK22}. Example depth maps are displayed in Figure ~\ref{fig:depth_example}. We also include a trivial constant-depth baseline. In this case, the 3D distance degenerates to the planar 2D distance on the image plane, allowing us to examine whether the gain comes merely from 2D local spatial proximity. 

Table~\ref{depth_quality} reports COCO object detection results after 200-epoch COCO pretraining. The constant-depth variant only brings a small improvement over MoCo v2, suggesting that simple 2D spatial proximity provides limited benefit. In contrast, using estimated depth substantially improves the baseline. Even the lightweight MiDaS Small$_{256}$ model improves $\text{AP}^{b}$ by \textbf{2.2}, showing that DGCL does not require highly accurate depth to be effective. Higher-quality depth generally leads to better performance, but the improvement saturates quickly. For example, stronger MiDaS models improve $\text{AP}^{b}$ to around \textbf{39.0} to \textbf{39.4}, with relatively small differences among $\text{Large}_{384}$, $\text{DPT}_{L-384}$, $\text{Swin2}_{L-384}$, and $\text{BEiT}_{L-512}$. This behavior is reasonable because DGCL is applied on feature maps, which is usually low-resolution. With a ResNet-50 backbone, a $224\times224$ input image is reduced to a $7\times7$ feature map, so fine-grained depth details are largely removed during downsampling. Therefore, coarse but reliable depth is already sufficient to provide useful relative 3D proximity for DGCL.

\begin{table}[H]
  \centering
   
  \begin{tabular}{@{}l|c|c @{}}
    \toprule
    \multirow{2}*{Method} & \multirow{2}*{\makecell{MiDaS\\depth quality}} &COCO Detection\\
    & & $\text{AP}^\text{b}$ / $\text{AP}_\text{50}^\text{b}$ / $\text{AP}_\text{75}^\text{b}$\\
    \midrule
    MoCo v2 & - & 36.5 / 56.0 / 39.7 \\
    \midrule
    Trivial  & - & 37.0 / 56.5 / 40.3\\
    Small\scriptsize{256}  & -76 & 38.7 / 58.4 / 42.2\\
    Large\scriptsize{384} &-32 & 39.1 / \textbf{59.2} / 43.1\\
    DPT\scriptsize{L-384} & 0 & 39.0 / 58.9 / 42.7\\
    Swin2\scriptsize{L-384} & +22 & \textbf{39.4} / 59.0 / 42.9\\
    BEiT\scriptsize{L-512} & +34 & 39.3 / 59.0 / \textbf{43.1}\\
    \bottomrule
  \end{tabular}
  \caption{COCO object detection performance with DGCL pre-training on COCO for 200 epochs using depth of different quality by MiDaS~\cite{DBLP:journals/pami/RanftlLHSK22}. DGCL substantially improves the baseline with only coarse depth estimated by a lightweight model, e.g., Small\scriptsize{256}.}
  \label{depth_quality}
\end{table}


%% file: sec/6_visualization.tex
\section{Visualizations}

\begin{figure}[H]
  \centering
  \newcommand{\visimg}[1]{%
    \begin{minipage}[t]{0.15\linewidth}
      \centering
      \includegraphics[width=\linewidth]{#1}
    \end{minipage}
  }
  \newcommand{\visimgcap}[2]{%
    \begin{minipage}[t]{0.15\linewidth}
      \centering
      \includegraphics[width=\linewidth]{#1}\\[-0.2em]
      {\scriptsize #2}
    \end{minipage}
  }

  \centering
  \begin{minipage}{1.0\linewidth}
  \visimg{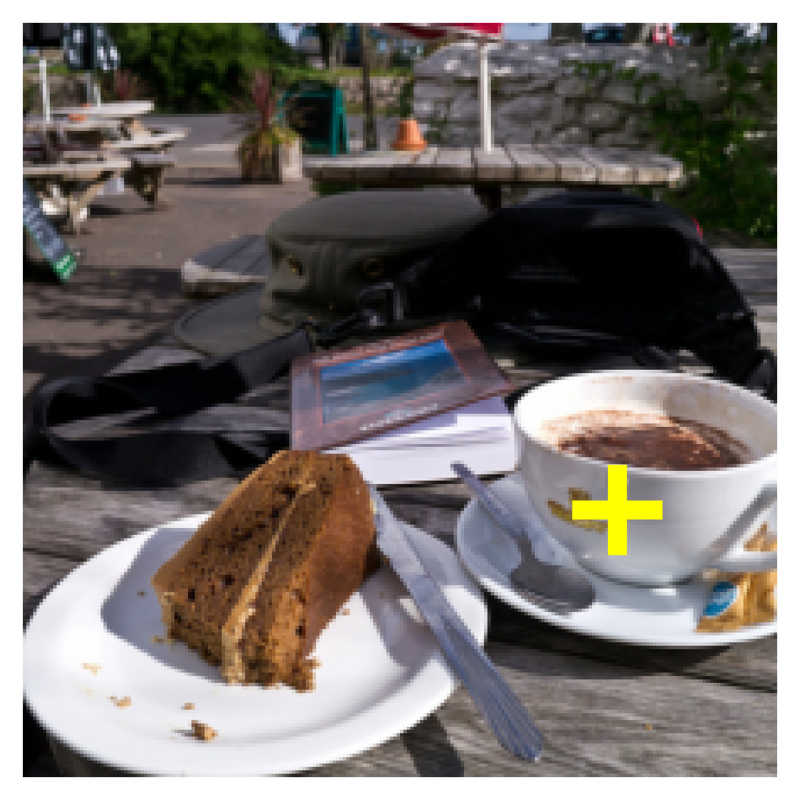}\hfill
  \visimg{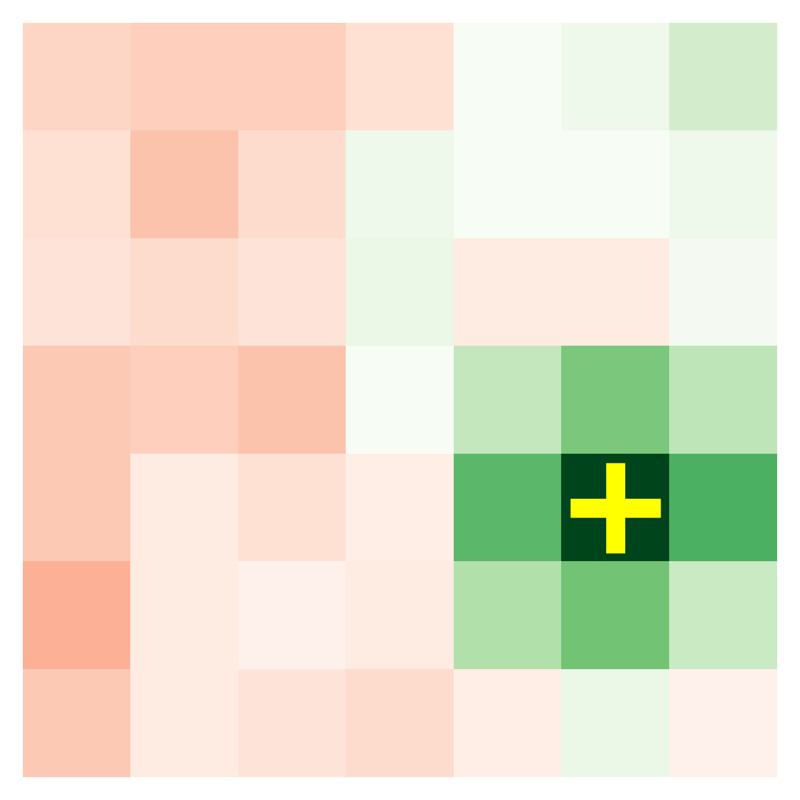}\hfill
  \visimg{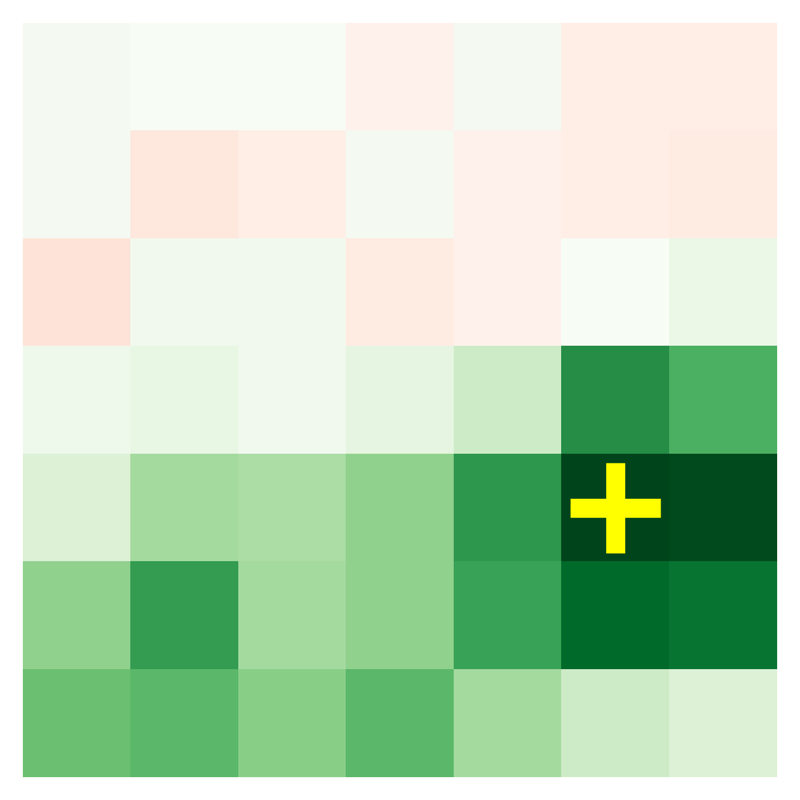}\hfill
  \visimg{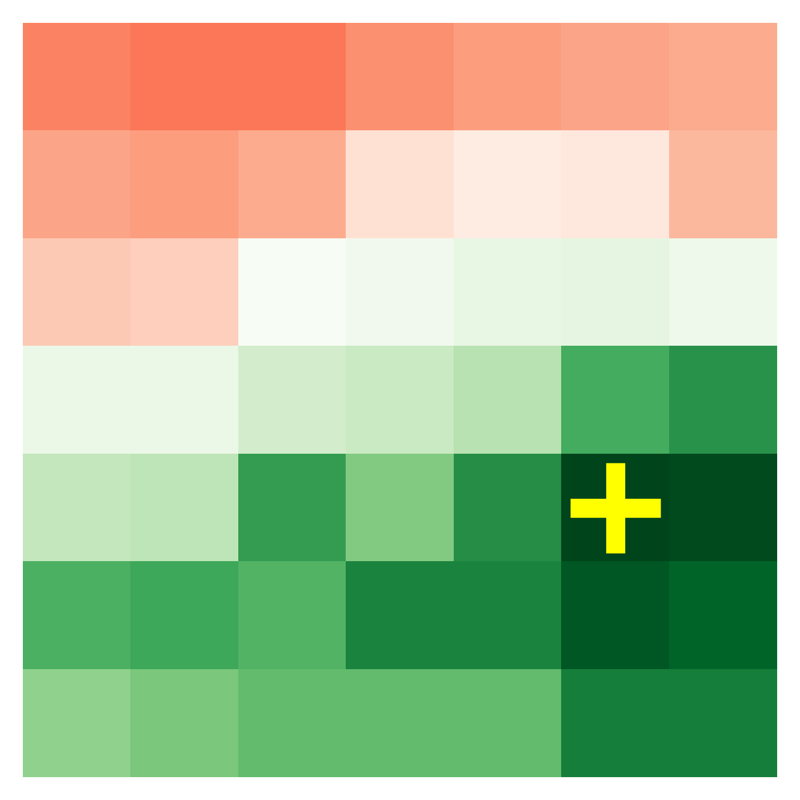}\hfill
  \visimg{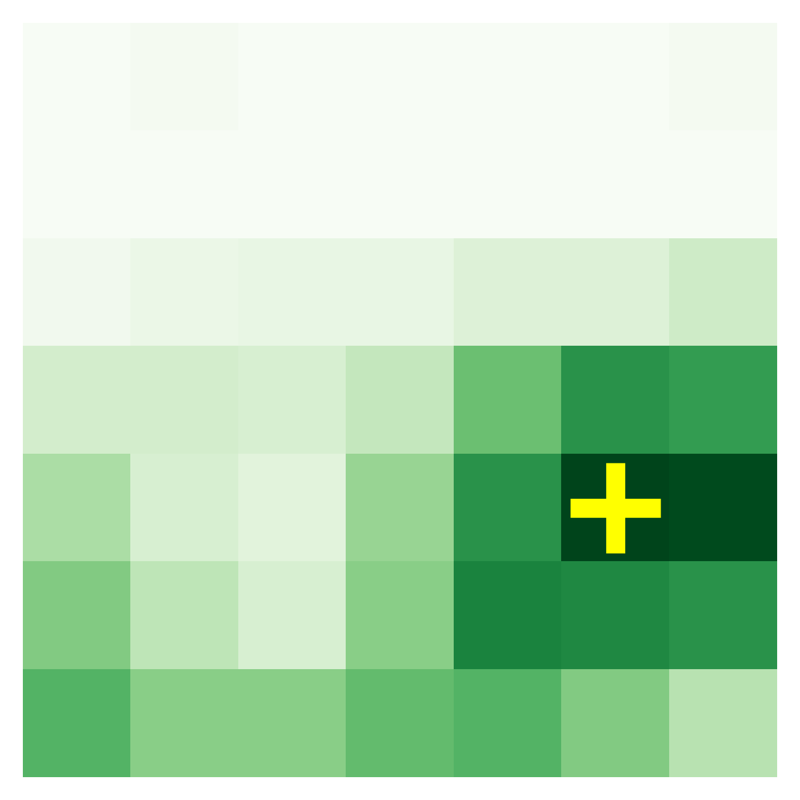}\hfill
  \visimg{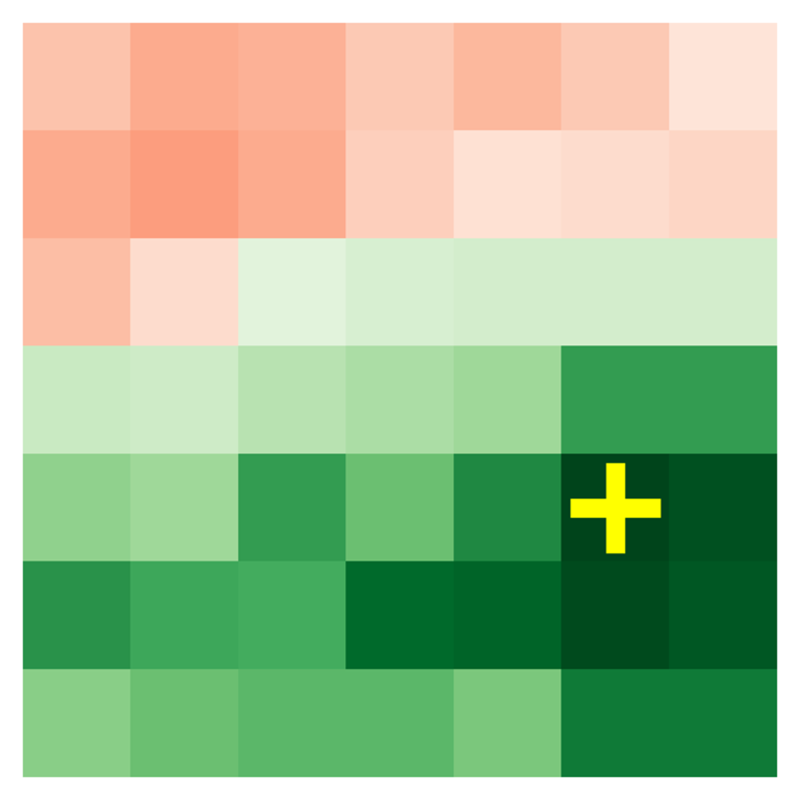}


  \visimg{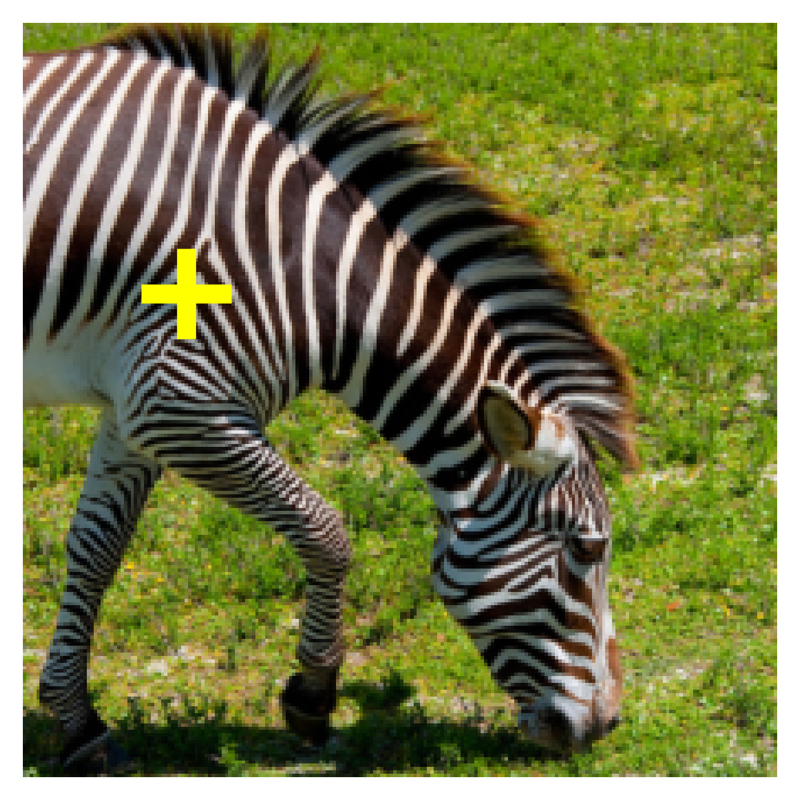}\hfill
  \visimg{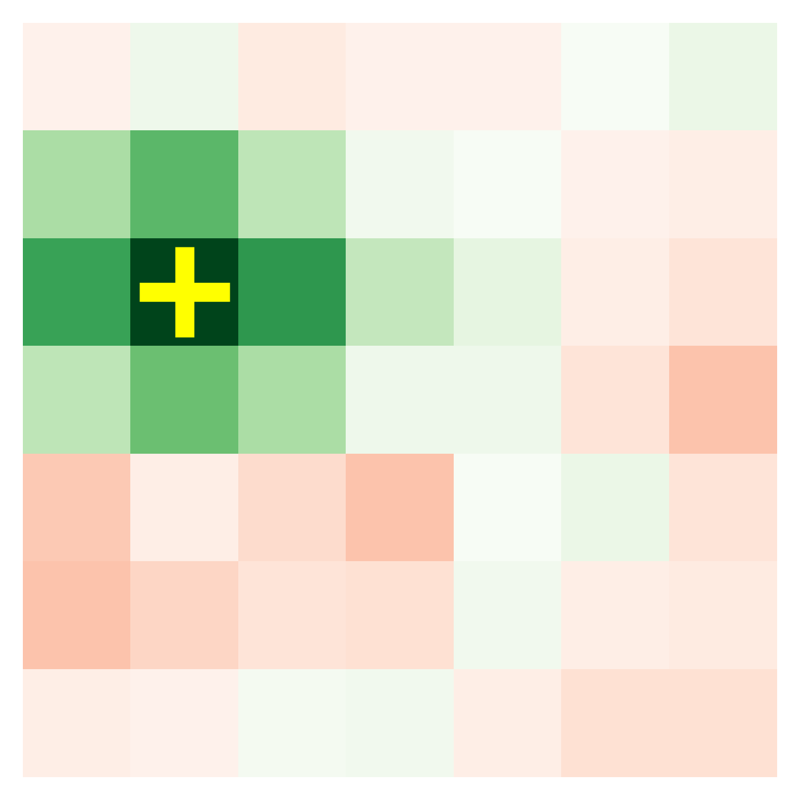}\hfill
  \visimg{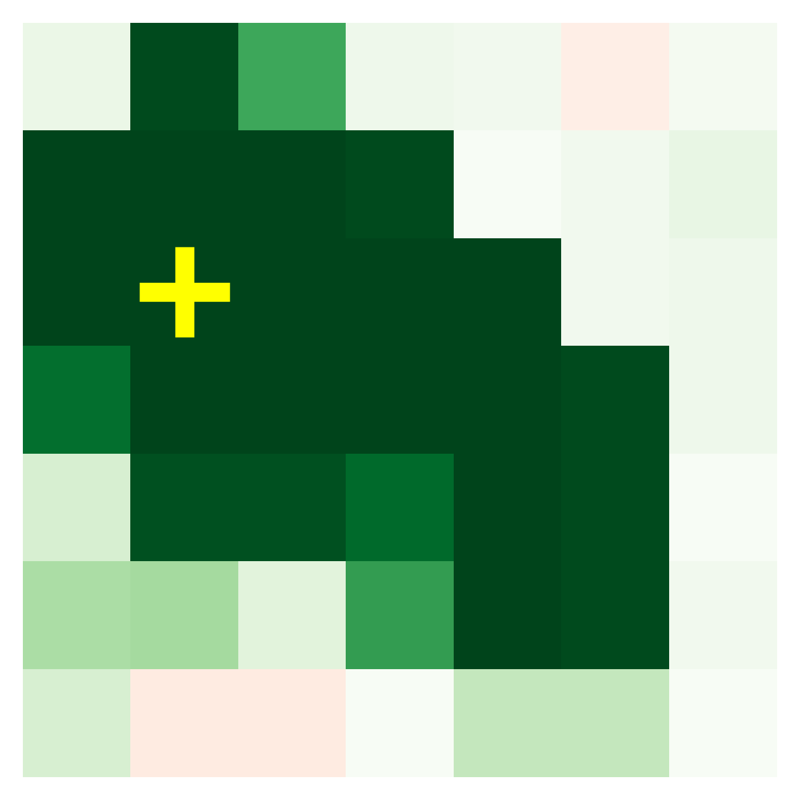}\hfill
  \visimg{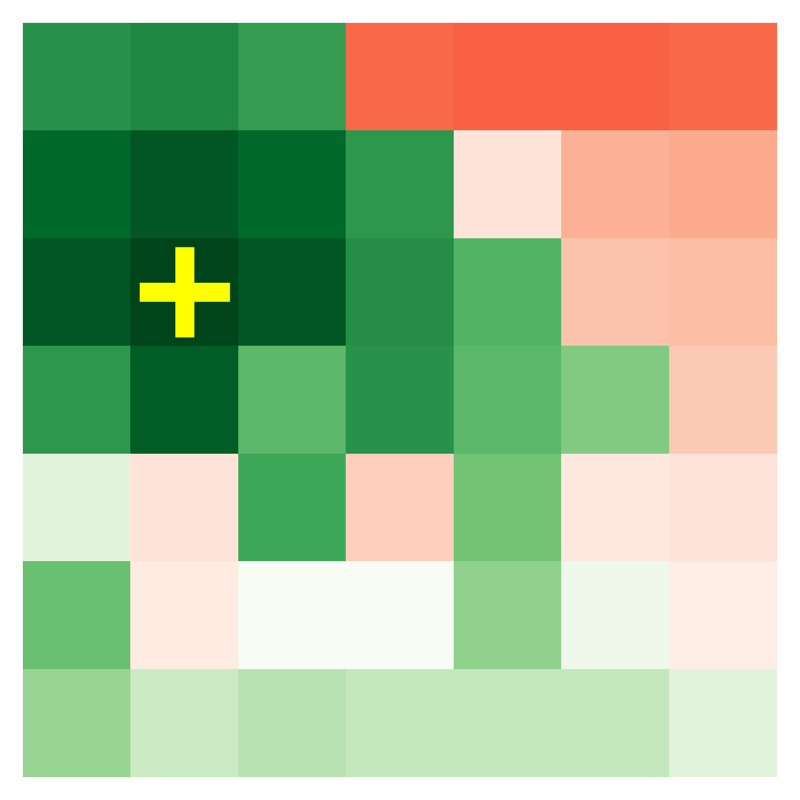}\hfill
  \visimg{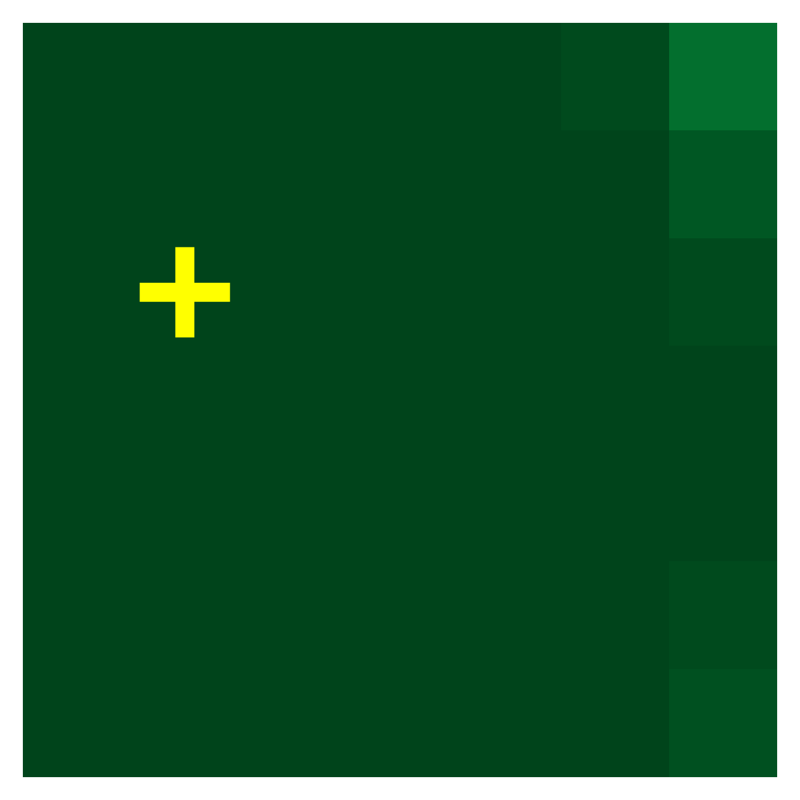}\hfill
  \visimg{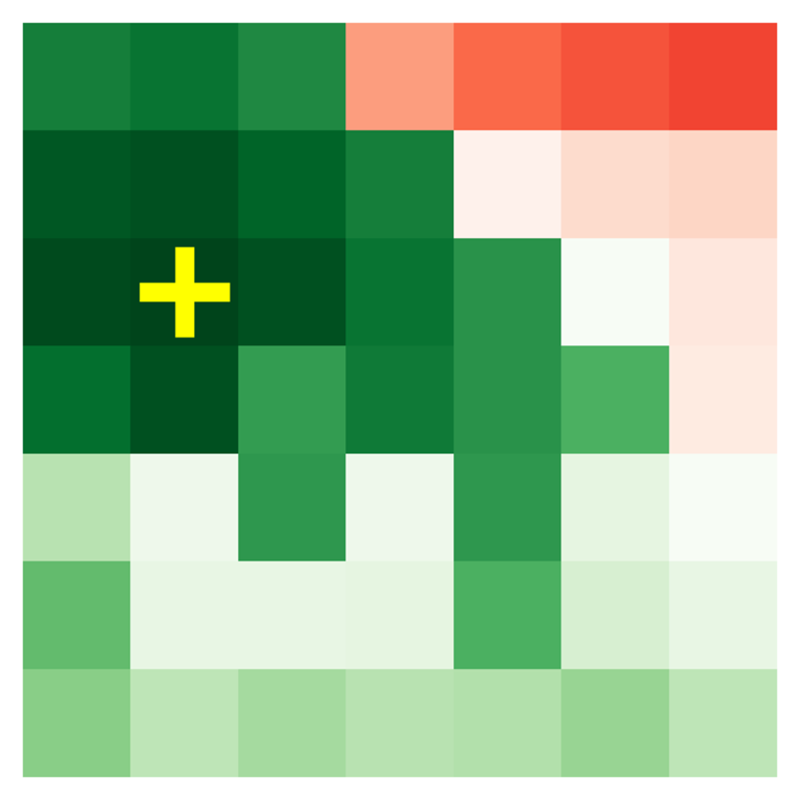}


  \visimgcap{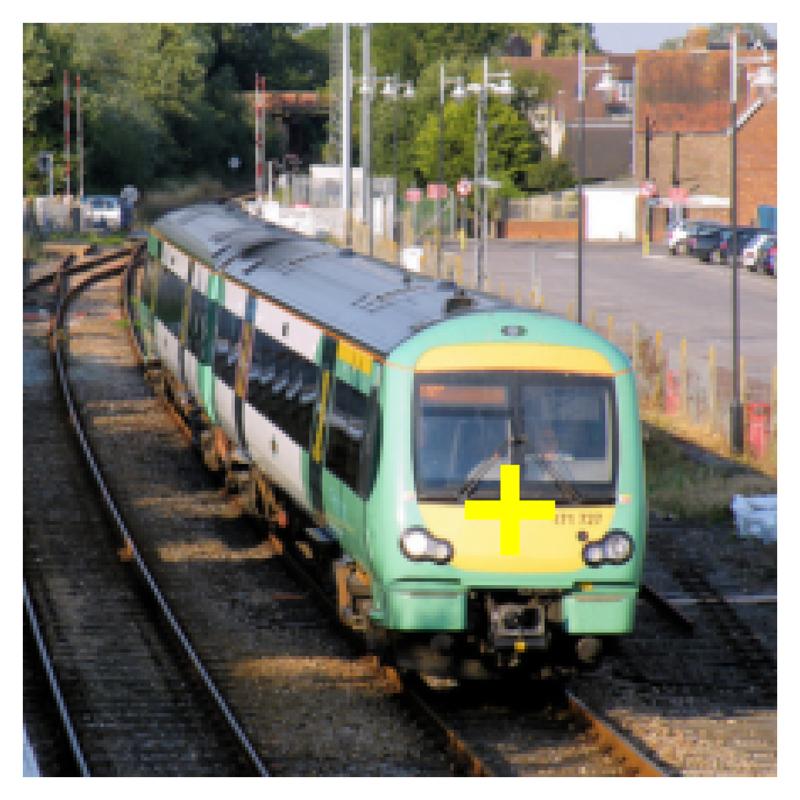}{Image}\hfill
  \visimgcap{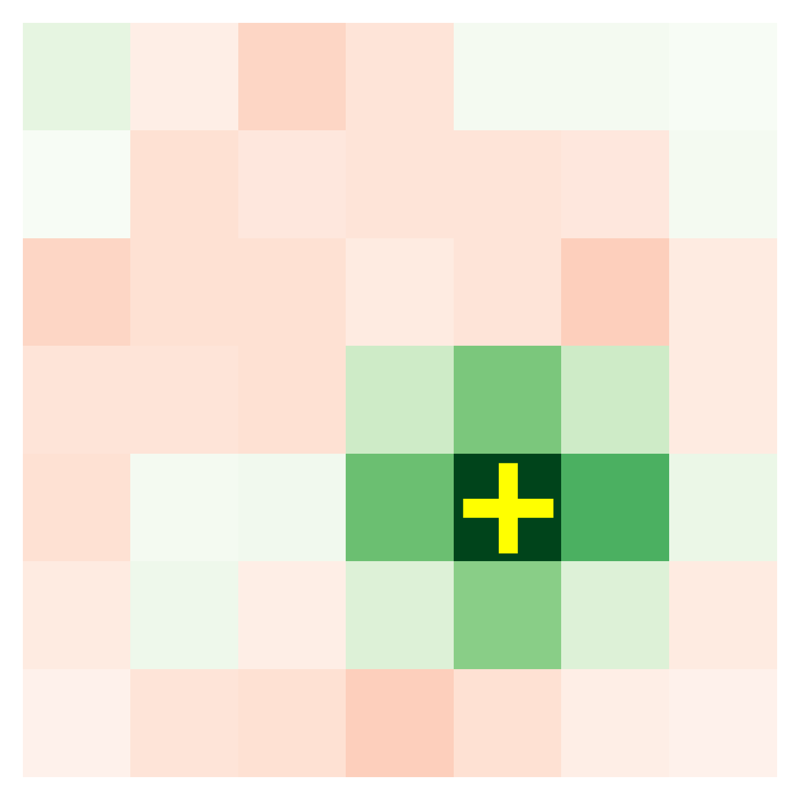}{PixPro~\cite{DBLP:conf/cvpr/XieL00L021}}\hfill
  \visimgcap{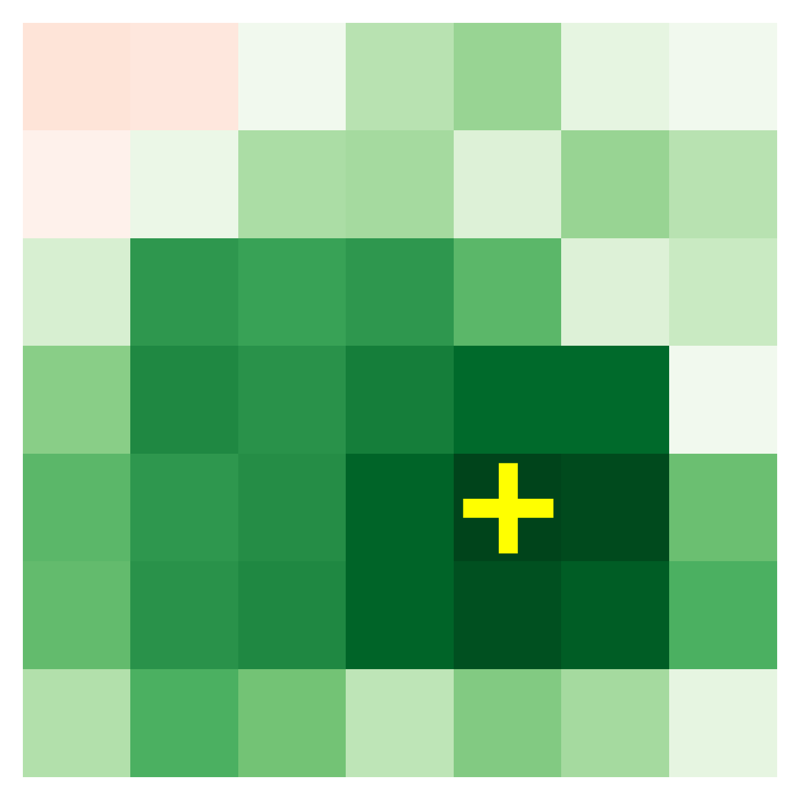}{MoCo v2~\cite{DBLP:journals/corr/abs-2003-04297}}\hfill
  \visimgcap{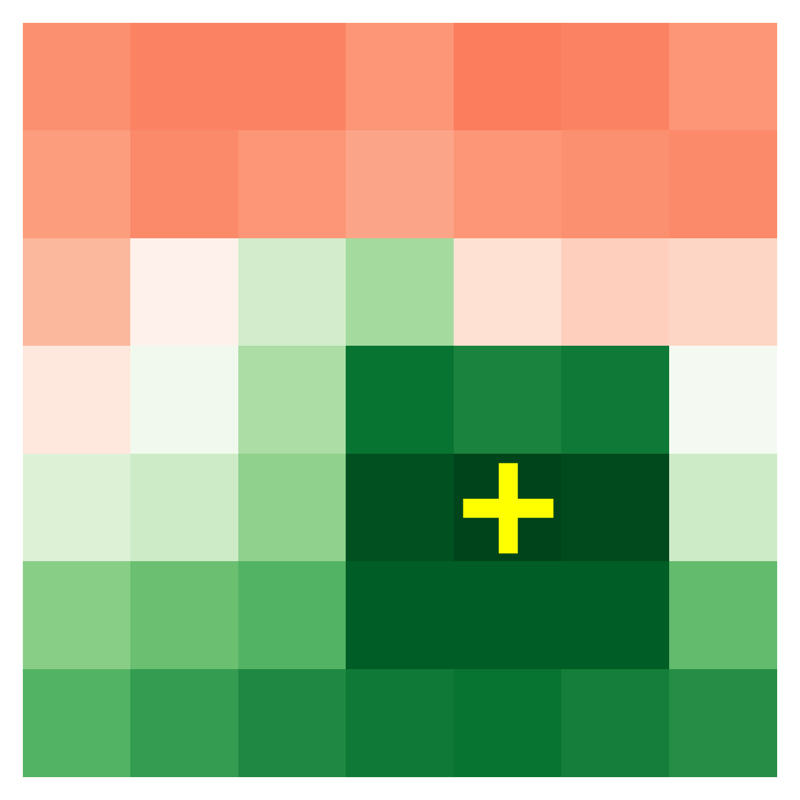}{DGCL-M2}\hfill
  \visimgcap{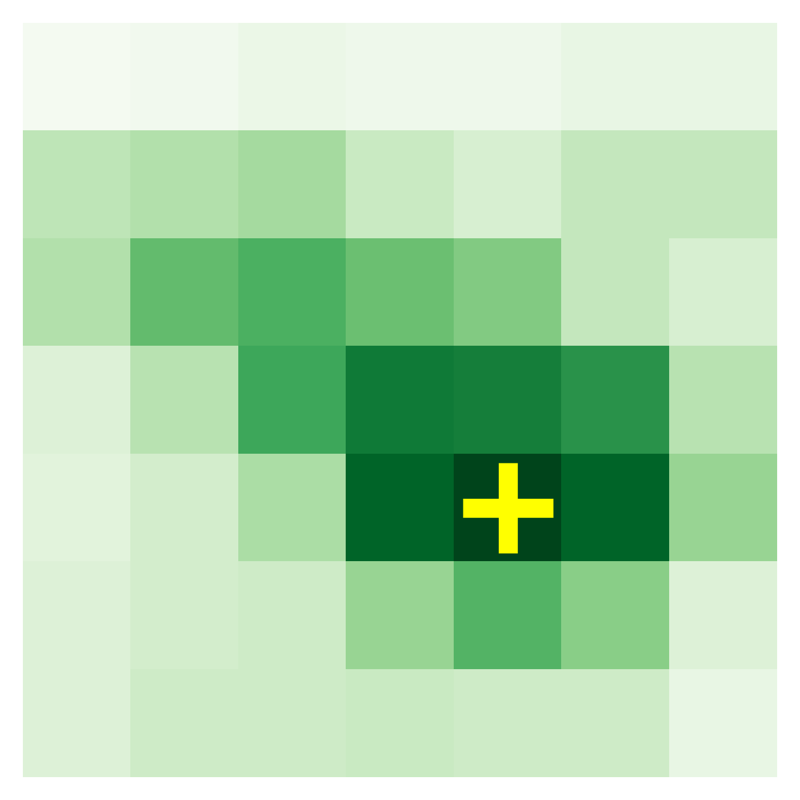}{SlotCon~\cite{DBLP:conf/nips/WenZZZQ22}}\hfill
  \visimgcap{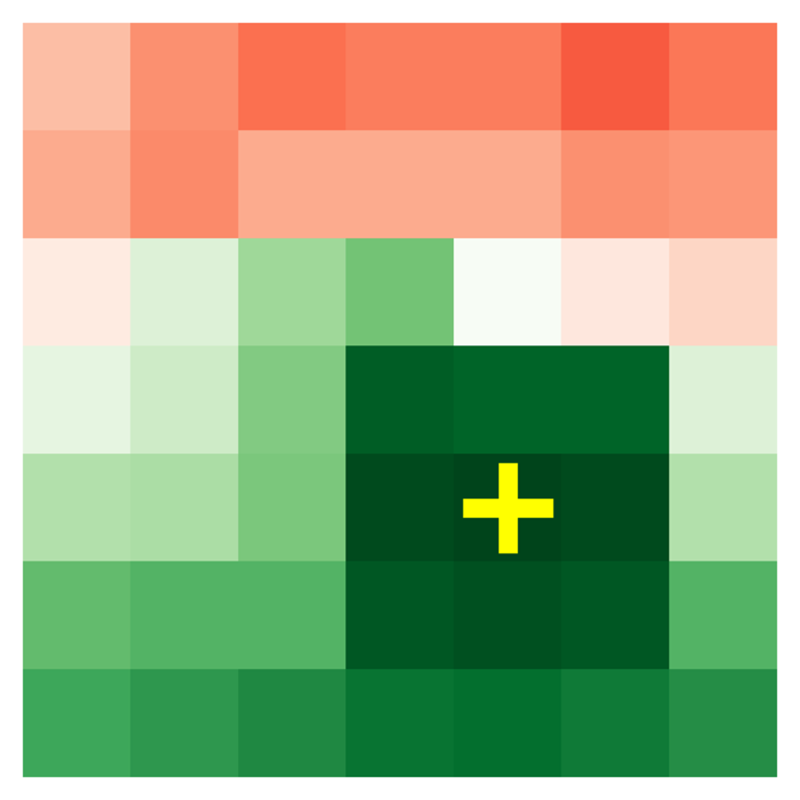}{DGCL-S}
  \end{minipage}


  \caption{Visualization of dense feature similarity with respect to the query pixel marked by the yellow cross. Green indicates higher similarity and red indicates lower similarity. Compared with other contrastive baselines, DGCL produces similarity maps that better align with geometrically and semantically related regions.}
  \label{fig:vis}
\end{figure}

We visualize dense representations in Figure~\ref{fig:vis} by computing the feature similarity between each pixel and a query pixel marked by the yellow cross. PixPro~\cite{DBLP:conf/cvpr/XieL00L021} tends to produce locally smooth similarity maps, reflecting its emphasis on spatially adjacent pixels by feature propagation. However, the resulting similarity may spread to regions that are close in the 2D image plane but less related in scene structure. MoCo v2~\cite{DBLP:journals/corr/abs-2003-04297}, which is trained mainly with image-level contrast, produces less localized dense similarity. SlotCon~\cite{DBLP:conf/nips/WenZZZQ22} learns more structured regions through data-driven grouping, but the grouping can sometimes become overly coarse or fail to align with meaningful object parts, and the resulting similarity maps can become less discriminative. In contrast, both DGCL-M2 and DGCL-S produce similarity maps that are more concentrated around geometrically and semantically related regions.

\section{Conclusions}

We presented Depth-Guided Contrastive Learning (DGCL), a simple auxiliary objective for injecting 3D spatial awareness into 2D contrastive representation learning. Unlike previous pixel-level contrastive methods that mainly define similarity from semantic perspectives, DGCL derives supervision from depth by assuming that pixels close in 3D space should have correlated representations. To avoid dependence on absolute depth scale, DGCL transforms 3D proximity into contrastive similarity through relative distance comparisons among randomly sampled pixels. This design makes DGCL easy to integrate with existing contrastive frameworks and applicable to depth maps from different sources, including estimated, synthetic, or sensor-provided depth. Extensive experiments across different datasets and models show that DGCL helps 2D representations explicitly capture 3D structure and consistently improves downstream transfer performance. Future work will explore how to better exploit fine-grained depth cues and higher-level 3D scene relationships, especially as stronger depth estimators continue to emerge~\cite{DBLP:conf/cvpr/YangKHXFZ24}. We hope our work can inspire the research of representation learning with 3D awareness.

%% file: sec/7_sup.tex
\clearpage
\setcounter{page}{1}
\maketitlesupplementary

\setcounter{section}{0}


\section{Hyperparameter Analysis}
\label{hyperparameter}

\subsection{Number of anchors \texorpdfstring{$K$}{K}}
\label{hyperparameter_k}
\begin{figure}[H]
  \centering
  \begin{minipage}{0.85\linewidth}
  \centering
  \begin{minipage}[t]{0.32\linewidth}
    \centering
    \includegraphics[width=\linewidth]{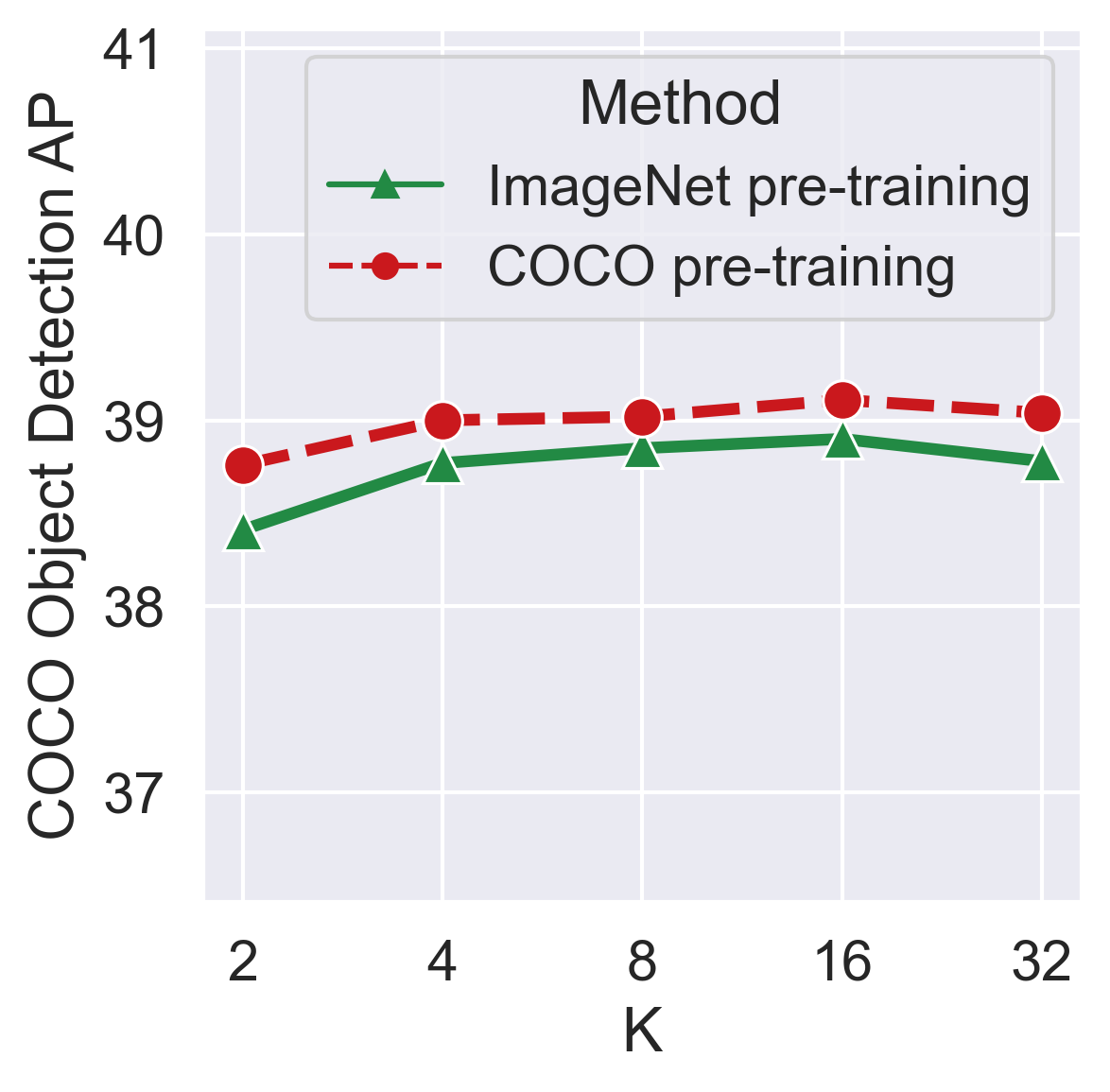}
  \end{minipage}
  \hfill
  \begin{minipage}[t]{0.32\linewidth}
    \centering
    \includegraphics[width=\linewidth]{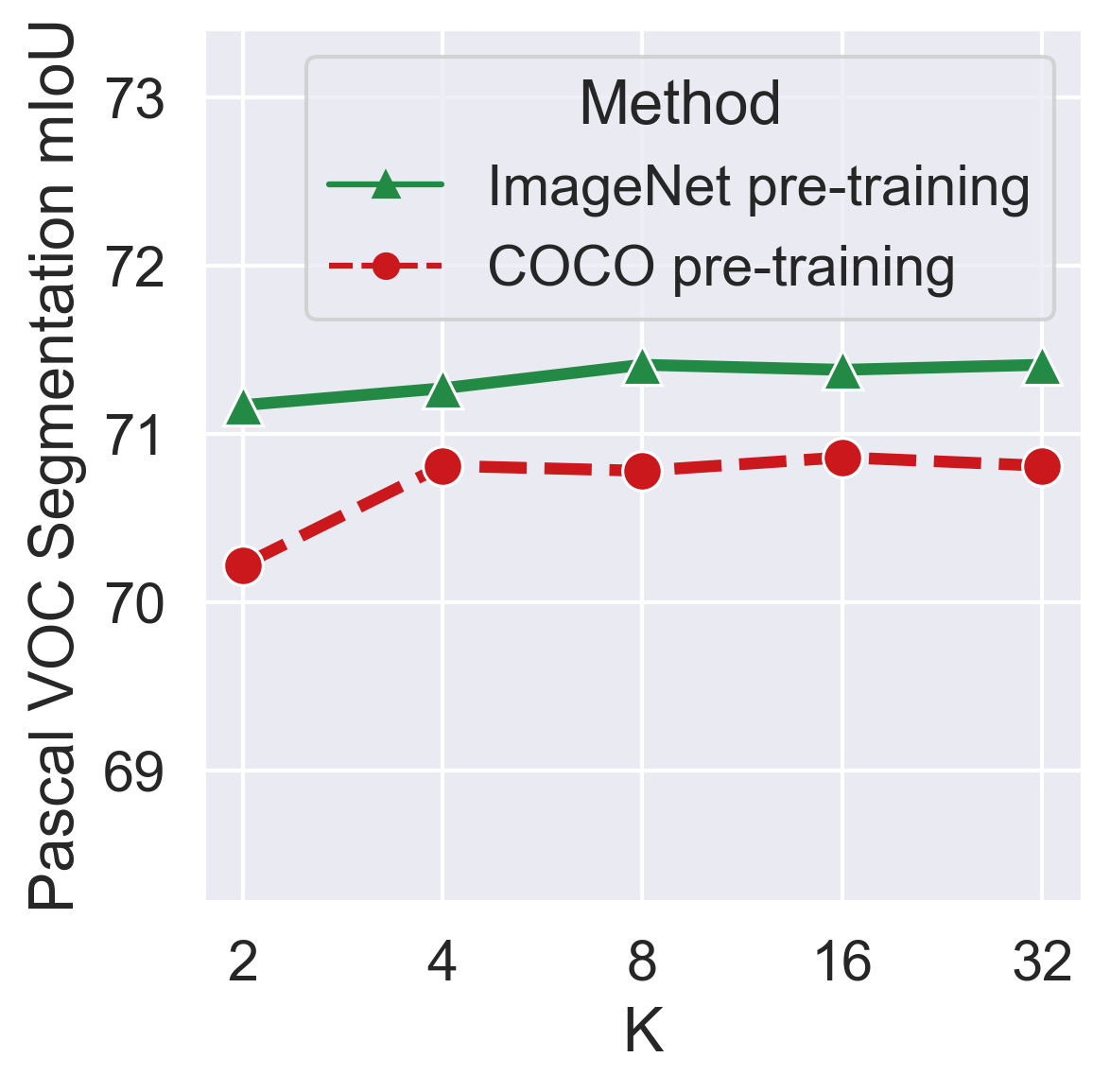}
  \end{minipage}
  \hfill
  \begin{minipage}[t]{0.32\linewidth}
    \centering
    \includegraphics[width=\linewidth]{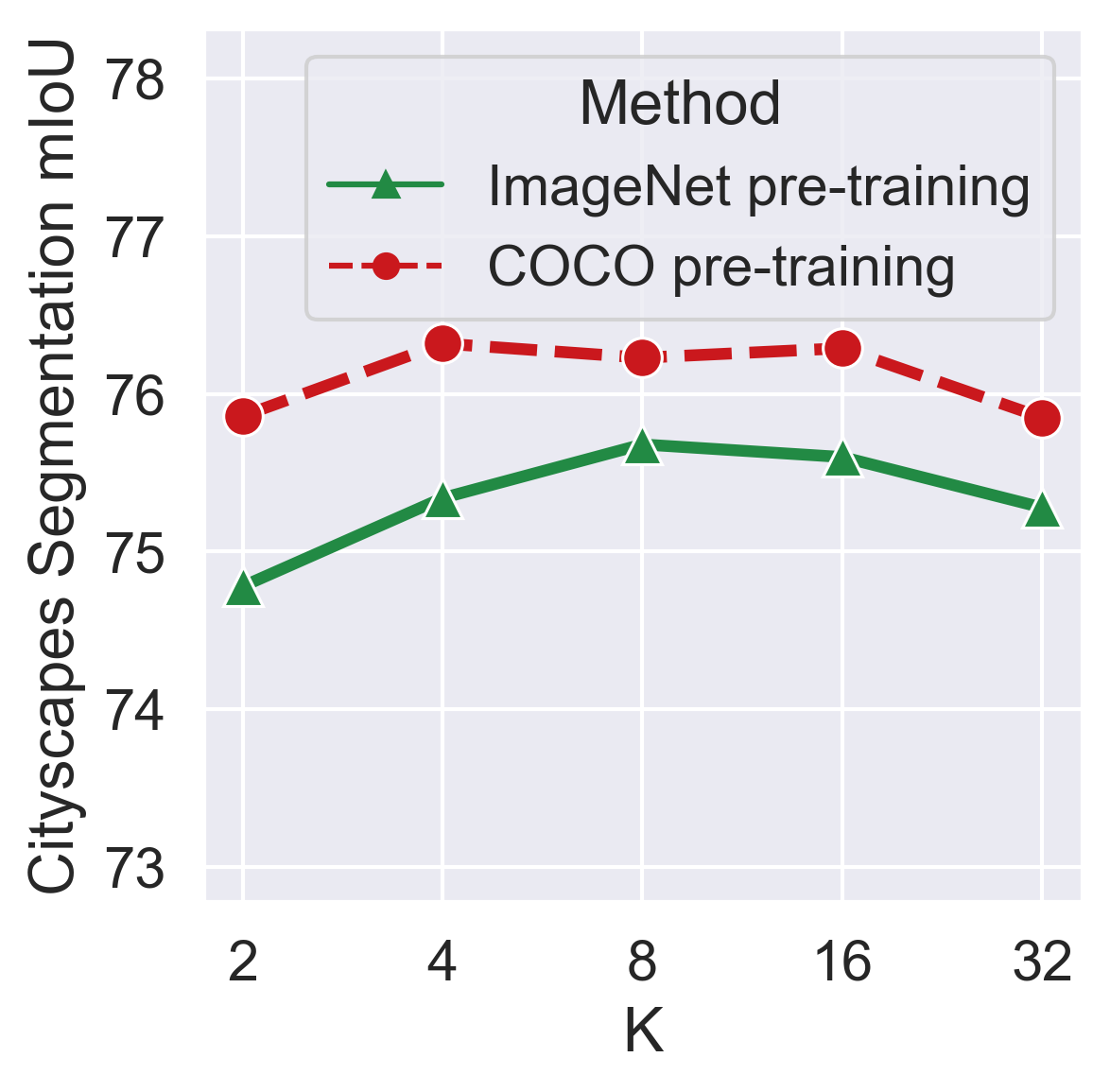}
  \end{minipage}
  \end{minipage}
  \par
  \caption{The best performance is consistently obtained around $K=8$ and insensitive to the semantic granularity of either pre-training data or downstream tasks, indicating the improvement is due to the learned 3D spatial awareness from 2D instead of approximating specific semantic labels by tuning $K$.}
  \label{fig:ablation_k}
\end{figure}

As discussed in Section~\ref{sec:methodology}, the number of anchors $K$ controls how concentrated the induced similarity is around each query pixel. A natural concern is that DGCL may require a carefully tuned $K$ to match the semantic granularity of the pretraining data or downstream tasks. We therefore study the effect of $K$ by pretraining DGCL with MoCo v2 on COCO~\cite{DBLP:conf/eccv/LinMBHPRDZ14}, which contains scene-centric images, and on a 118k-image subset of ImageNet~\cite{ILSVRC15}, which is more object-centric. The pretrained models are then transferred to downstream tasks with different semantic granularity, including COCO object detection, Pascal VOC semantic segmentation~\cite{DBLP:journals/ijcv/EveringhamGWWZ10}, and Cityscapes semantic segmentation~\cite{DBLP:conf/cvpr/CordtsORREBFRS16}. As shown in Figure~\ref{fig:ablation_k}, DGCL is relatively stable across a range of $K$, with the best performance consistently obtained around $K=8$. Very small or very large values lead to mild performance drops. When $K$ is too small, the nearest-anchor prediction task becomes less informative because the model only compares among very few candidates. When $K$ is too large, the sampled anchors are more likely to contain pixels extremely close to the query, making the task easier and less effective. 

\subsection{Number of anchor groups \texorpdfstring{$N$}{N}}
\label{hyperparameter_n}
\begin{figure}[H]

  \centering
  \begin{minipage}{1.0\linewidth}
  \centering
  \begin{minipage}[b]{0.28\linewidth}
    \centering

    \begin{minipage}[b]{0.48\linewidth}
      \centering
      \includegraphics[width=\linewidth]{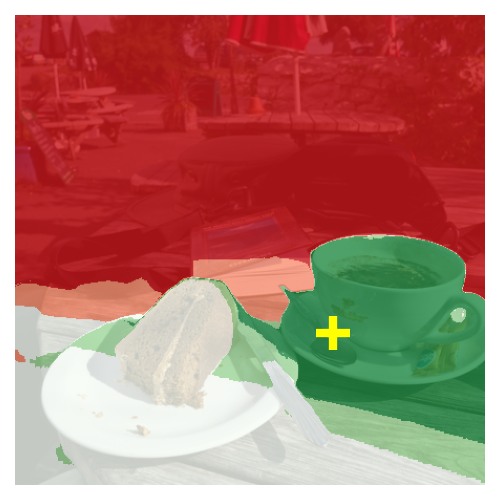}\\
      {\scriptsize $N=4$}
    \end{minipage}
    \hfill
    \begin{minipage}[b]{0.48\linewidth}
      \centering
      \includegraphics[width=\linewidth]{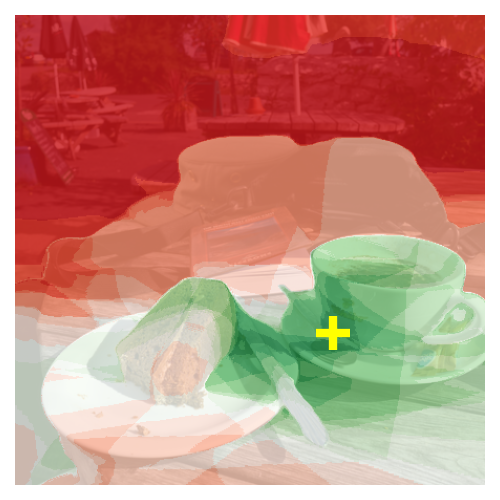}\\
      {\scriptsize $N=16$}
    \end{minipage}
    \begin{minipage}[b]{0.48\linewidth}
      \centering
      \includegraphics[width=\linewidth]{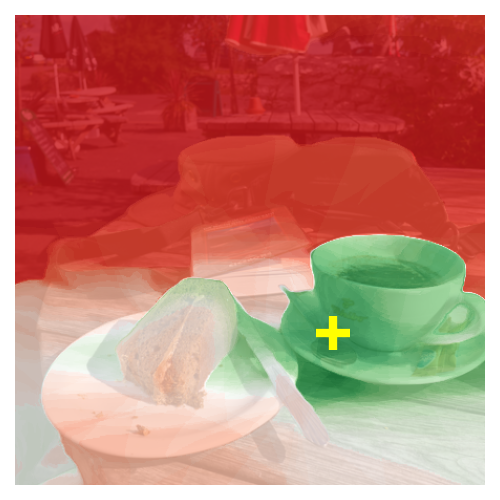}\\
      {\scriptsize $N=64$}
    \end{minipage}
    \hfill
    \begin{minipage}[b]{0.48\linewidth}
      \centering
      \includegraphics[width=\linewidth]{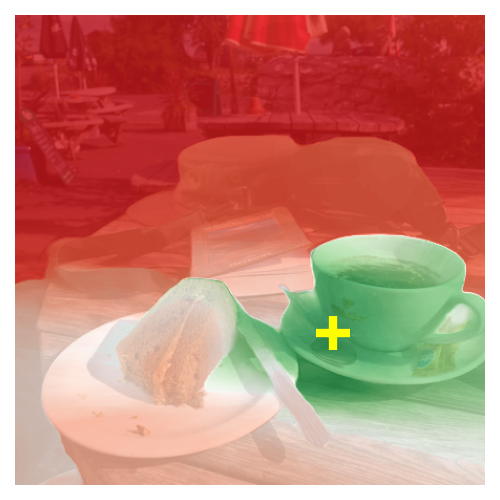}\\
      {\scriptsize $N=256$}
    \end{minipage}

  \end{minipage}
  \hfill
  \begin{minipage}[b]{0.32\linewidth}
    \centering
    \includegraphics[width=\linewidth]{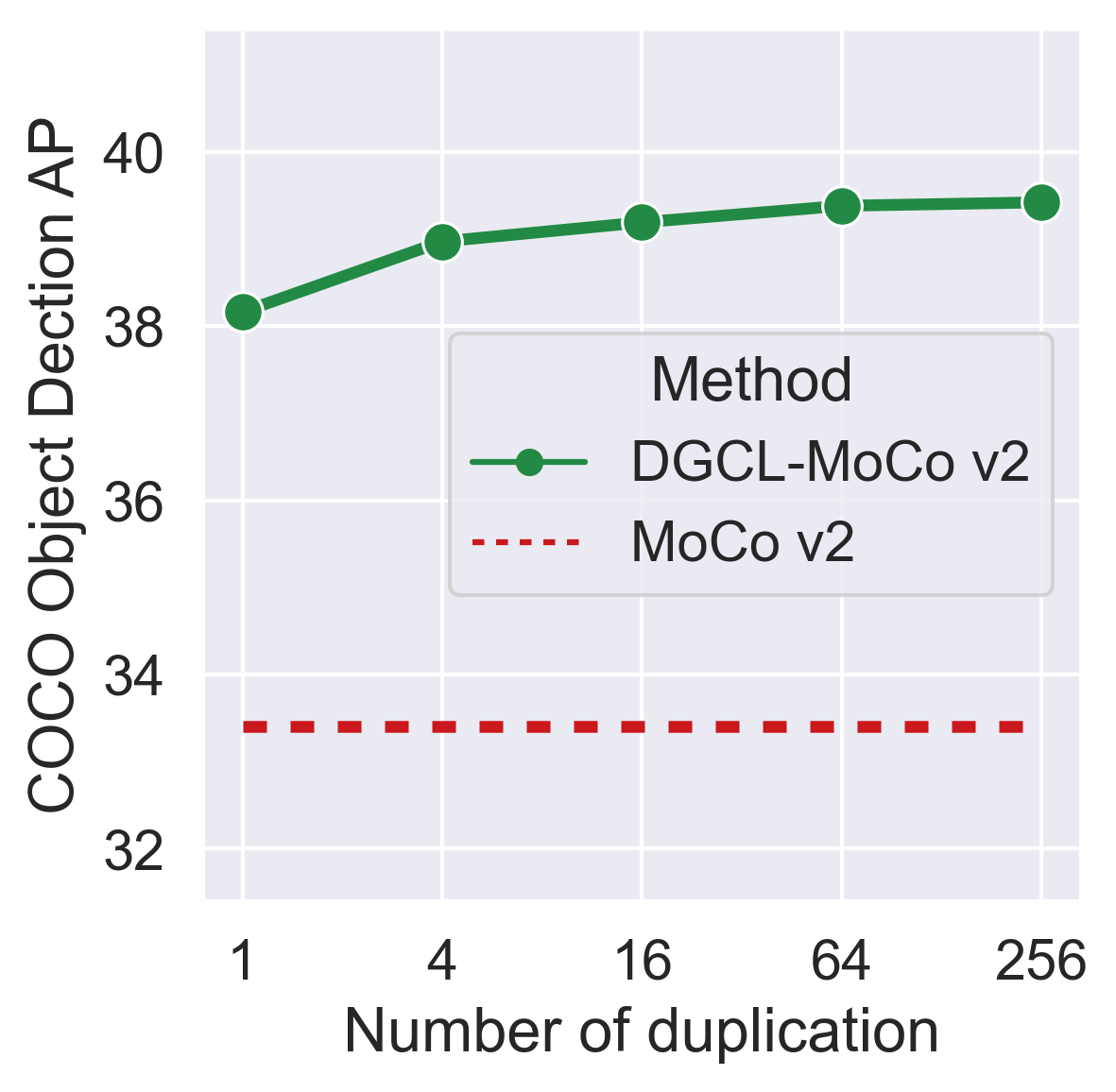}
  \end{minipage}
  \hfill
  \begin{minipage}[b]{0.34\linewidth}
    \centering
    \includegraphics[width=\linewidth]{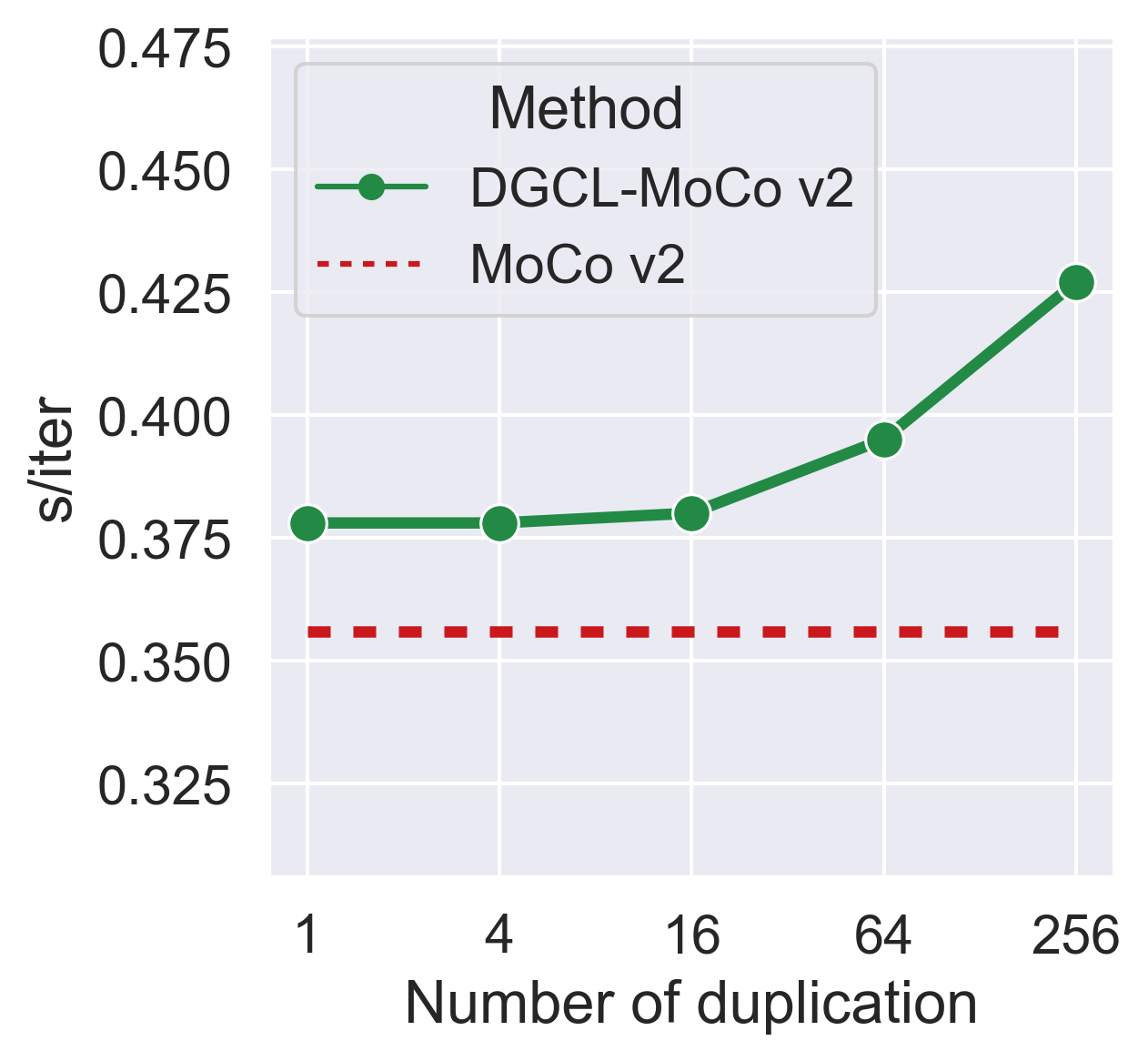}
  \end{minipage}
  \end{minipage}
  \par

  \caption{Increasing the number of anchor groups $N$ provides a smoother approximation of the 3D similarity distribution for learning in each iteration and improve the performance. The performance plateau is reached by 64 groups with only a 10\% additional overhead, measured on 2 RTX 2080 Ti GPUs.}
  \label{fig:ablation_n}
\end{figure}

The number of anchor groups $N$ controls how many independent nearest-anchor prediction tasks are constructed for each image in one iteration. A larger $N$ samples more anchor groups and therefore provides a smoother approximation of the rank-based 3D similarity distribution. As shown in Figure~\ref{fig:ablation_n}, increasing $N$ improves transfer performance at first, and the performance saturates around $N=64$. We also measure the computational overhead of increasing $N$ using ResNet-50-based MoCo v2 with $224\times224$ inputs on two RTX 2080 Ti GPUs. Using $N=64$ reaches the performance plateau while increasing the iteration time by only about 10\%. It is also worth noting that additional memory usage is almost unnoticeable with $N$ smaller than 256. Therefore, we use $N=64$ as the default setting in our experiments, as it provides a good trade-off between accuracy and efficiency. 

\section{Convergence}
\label{Convergence}
\begin{figure}[H]
  \centering
  \begin{subfigure}{0.48\linewidth}
    \includegraphics[width=1\linewidth]{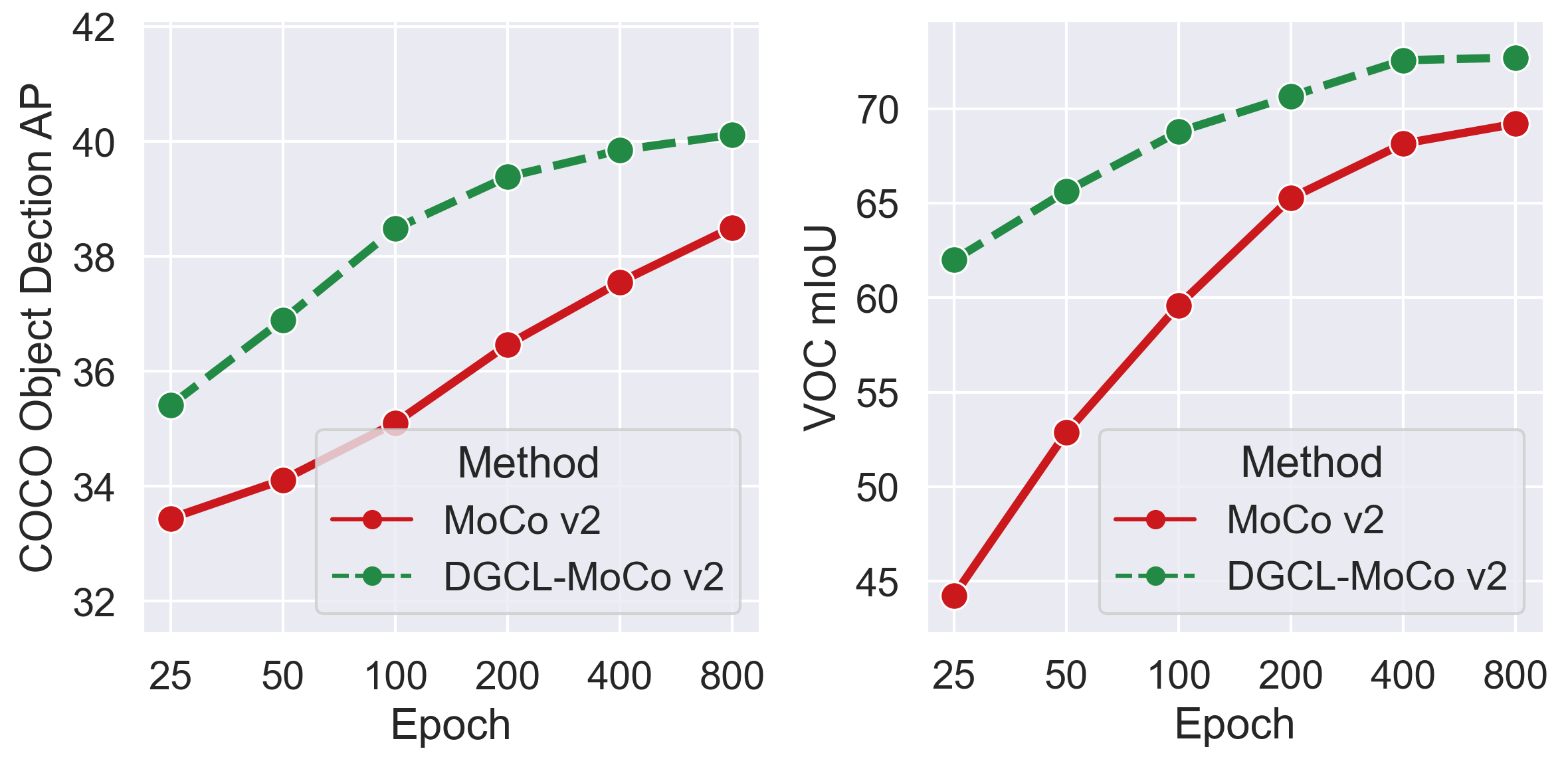}
    \caption{MoCo v2 and DGCL-MoCo v2 pre-training}
    \label{moco}
  \end{subfigure}
  \hfill
  \begin{subfigure}{0.48\linewidth}
    \includegraphics[width=1\linewidth]{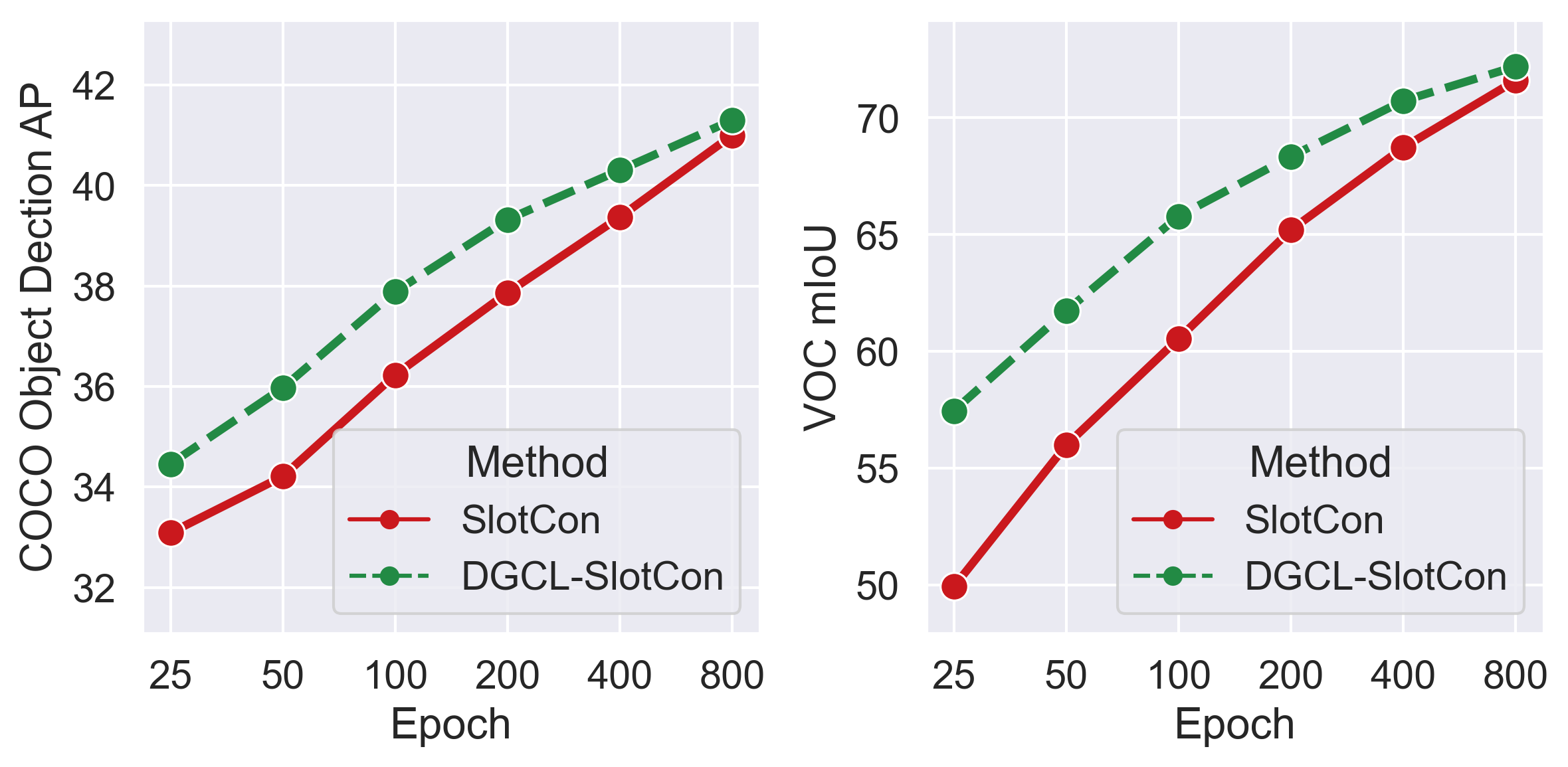}
    \caption{SlotCon and DGCL-SlotCon pre-training}
    \label{slotcon}
  \end{subfigure}
  \caption{Convergence comparison between DGCL and the corresponding base contrastive frameworks. DGCL improves both MoCo v2 and SlotCon across different pretraining epochs, especially in the early training stage, indicating that DGCL accelerates the learning of transferable dense representations.}
  \label{epoch}
\end{figure}

As shown in Figure~\ref{epoch}, DGCL improves convergence for both MoCo v2 and SlotCon. The advantage is especially clear in the early training stage, where training with DGCL achieves higher COCO detection AP and VOC mIoU than the corresponding baseline under the same number of pretraining epochs. It indicates that DGCL provides an effective additional learning signal, helping the model discriminate image contents and then learn transferable dense representations more efficiently. Such faster convergence is particularly useful when the pretraining budget is limited. 

\section{Standard Deviations}
\label{standard deviation}
To assess the robustness of the improvements, we report the mean and standard deviation over three independent runs in Table~\ref{stat_result_im} and ~\ref{stat_result_cc} corresponding to the main results in Table ~\ref{im_result} and ~\ref{coco_result}. The small standard deviations indicate that the results are stable across runs, supporting the consistency of the gains reported in the main experiments.

\begin{table}[h]
  \centering
  \begin{tabular}{@{}l|c|c @{}}
    \toprule
     Metric    & DGCL-M2   & DGCL-S \\
    \midrule
     COCO  $\text{AP}^\text{b}$  & 40.2±0.1 & 42.1±0.1 \\
     COCO  $\text{AP}^\text{m}$  & 37.1±0.1 & 38.2±0.1 \\
     COCO  $\text{AP}^\text{k}$  & 66.6±0.2 & 67.2±0.2 \\
     City. mIoU&  76.4±0.2 & 77.0±0.1 \\
     VOC mIoU  &  76.2±0.2 & 76.9±0.2 \\
     ADE mIoU  &  38.7±0.2 & 40.3±0.2 \\
    \bottomrule
  \end{tabular}
  \caption{Standard deviation of downstream performance of DGCL models pretrained on ImageNet.}
  \label{stat_result_im}
\end{table}

\begin{table}[h]
  \centering
  \begin{tabular}{@{}l|c|c @{}}
    \toprule
     Metric    & DGCL-M2   & DGCL-S \\
    \midrule
     COCO  $\text{AP}^\text{b}$  & 40.1±0.1 & 41.3±0.1 \\
     COCO  $\text{AP}^\text{m}$  & 36.9±0.1 & 37.9±0.1 \\
     COCO  $\text{AP}^\text{k}$  & 66.7±0.2 & 67.4±0.1 \\
     City. mIoU&  76.4±0.3 & 76.7±0.2 \\
     VOC mIoU  &  72.4±0.3 & 72.7±0.2 \\
     ADE mIoU  &  38.9±0.2 & 39.9±0.2 \\
    \bottomrule
  \end{tabular}
  \caption{Standard deviation of downstream performance of DGCL models pretrained on COCO.}
  \label{stat_result_cc}
\end{table}

\section{Per-Category Performance Gain}

To further examine whether the benefits of learned 3D awareness are general or the 3D proximity coincide with certain semantic categories, we analyze the per-category improvement in COCO instance segmentation, with categories ordered by their average object size. The gains are observed for most categories across small and large objects and across diverse object shapes. This suggests that DGCL does not simply exploit object-size bias or approximate particular semantic classes, but provides generalizable 3D-aware cues that benefit semantic tasks.

Performance drops mainly occur for small or thin objects, such as stop signs and parking meters, or small objects in cluttered contexts, such as oranges. These cases may be sensitive to depth errors and downsampling, where object pixels may be mixed with nearby background or supporting surfaces. Thus, local 3D proximity can sometimes conflict with instance boundaries. This is a limitation of DGCL, but the overall per-category trend suggests that the geometric signal remains beneficial for most classes.

\begin{figure}[H]
  \centering
  \includegraphics[width=1.0\linewidth]{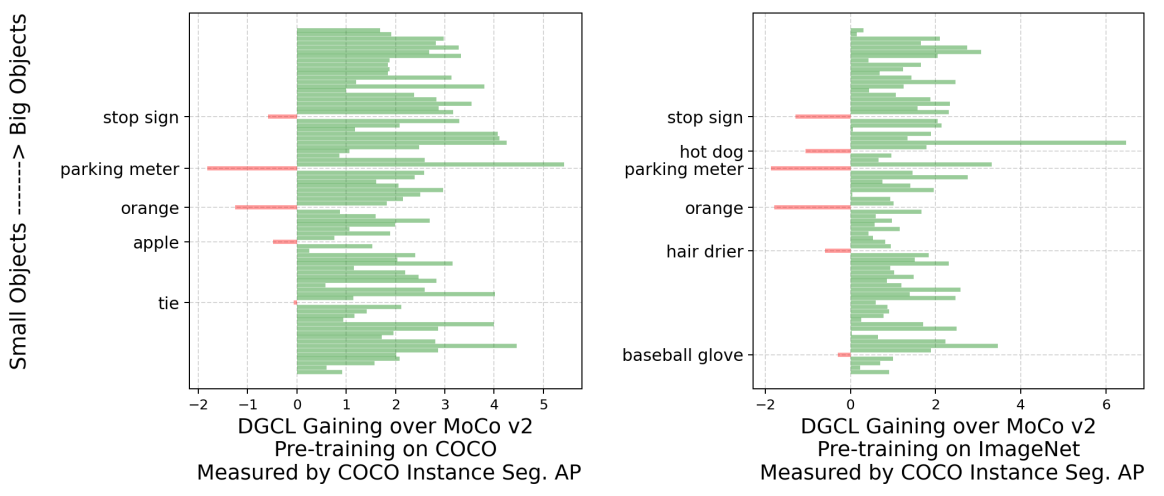}
  \caption{Per-category COCO instance segmentation gains of DGCL-M2 over MoCo v2, with categories ordered by average object size. DGCL improves most categories across small and large objects, suggesting that the learned 3D-aware representation provides generalizable geometric cues. Negative gains mainly appear for small, thin, or cluttered objects potentially due to depth errors and downsampling.}
\end{figure}